\documentclass[journal]{IEEEtran}

\usepackage{cite}
\usepackage{xcolor}
\definecolor{newgreen}{HTML}{228B22}
\usepackage[colorlinks,
            linkcolor=black,
            anchorcolor=black,
            citecolor=black,
            urlcolor=black
            ]{hyperref}
\usepackage{hyperref}
\usepackage{balance}

\ifCLASSINFOpdf
  \usepackage[pdftex]{graphicx}
\else
\fi
\usepackage{amsmath}
\usepackage{amssymb}
\usepackage{algorithm}
\usepackage{algorithmic}

\usepackage{array}
\usepackage{url}
\usepackage{makecell}
\usepackage{multirow}

\begin{document}
%
\title{Multitask Non-Autoregressive Model\\ for Human Motion Prediction}
%
%
%

\author{Bin~Li,
        Jian~Tian,
        Zhongfei~Zhang,
        Hailin~Feng,
        and~Xi~Li*
\thanks{Corresponding author: Xi Li}
\thanks{B. Li is with the College of Information Science and Electronic Engineering, Zhejiang University, Hangzhou 310027, China (e-mail: bin\_li@zju.edu.cn).}
\thanks{J. Tian and X. Li are with the College of Computer science and Technology, Zhejiang University, Hangzhou 310027, China (e-mail: \{tianjian29, xilizju\}@zju.edu.cn).}
\thanks{Z. Zhang is with the College of Information Science and Electronic Engineering, Zhejiang University, Hangzhou 310027, China, and also with the Computer Science Department, Watson School, The State University of New York, Binghamton, NY 13902 USA (e-mail: zhongfei@zju.edu.cn).}
\thanks{H. Feng is with School of Information Engineering, Zhejiang A\&F University, Hangzhou, 310027, China (email: hlfeng@zafu.edu.cn).}}

\maketitle

\begin{abstract}
  Human motion prediction, which aims at predicting future human skeletons given the past ones, is a typical sequence-to-sequence problem. 
  Therefore, extensive efforts have been continued on exploring different RNN-based encoder-decoder architectures. 
  However, by generating target poses conditioned on the previously generated ones, these models are prone to bringing issues such as error accumulation problem. 
  In this paper, we argue that such issue is mainly caused by adopting autoregressive manner.
  Hence, a novel Non-auToregressive Model (NAT) is proposed with a complete non-autoregressive decoding scheme, as well as a context encoder and a positional encoding module.
  More specifically, the context encoder embeds the given poses from temporal and spatial perspectives.
  The frame decoder is responsible for predicting each future pose independently.
  The positional encoding module injects positional signal into the model to indicate temporal order.
  %
  %
  Moreover, a multitask training paradigm is presented for both low-level human skeleton prediction and high-level human action recognition, resulting in the convincing improvement for the prediction task.
  Our approach is evaluated on Human3.6M and CMU-Mocap benchmarks and outperforms state-of-the-art autoregressive methods.
\end{abstract}


%
\IEEEpeerreviewmaketitle

\section{Introduction}
\label{sec:intro}
  \IEEEPARstart{A}{s } an important and challenging problem in computer vision, human motion prediction is typically formulated as a sequence modeling problem, which aims to predict a set of future human skeletons based on some existing real skeleton sequence data.
  Therefore, a natural solution to such a problem is to establish effective inertial motion models for capturing the temporal dependency among consecutive human skeleton frames~\cite{Lehrmann_2014_CVPR, wang2007gaussian, yao2011learning, taylor2007modeling, holden2016deep, Fragkiadaki_2015_ICCV, Jain_2016_CVPR, Martinez_2017_CVPR, Li_2018_CVPR, Guo_2019_human, Aksan_2019_ICCV, Liu_2019_CVPR, Mao_2019_ICCV, Pavllo_2018_BMVC, pavllo2019modeling, Gui_2018_ECCV, Hernandez_2019_ICCV}. 
  
  In general, these inertial motion models are based on either sequential autoregression or sequence-to-sequence encoder-decoder learning, which generates a sequence of future human skeletons in a recurrent frame-by-frame way, \emph{i.e.}, predicting the next generated frame depending on the existing real frames as well as the current generated frame.
  Usually, such inertial motion models are likely to face the following two challenges:
  1) Error accumulation: the prediction accuracy of the current frame relies heavily on that of previous frames, resulting in the recurrent prediction error propagation over time~\cite{zhou2018autoconditioned}.
  2) Mean pose problem: these models often converge to an undesired mean pose in the long-term predictions, \emph{i.e.}, the predictor gives rise to static predictions similar to the mean of the ground truth of future sequences~\cite{Li_2018_CVPR}.
  In this paper, we mainly focus on the error accumulation problem.

\begin{figure}[!t]
  \centering
    \includegraphics[width=0.9\linewidth]{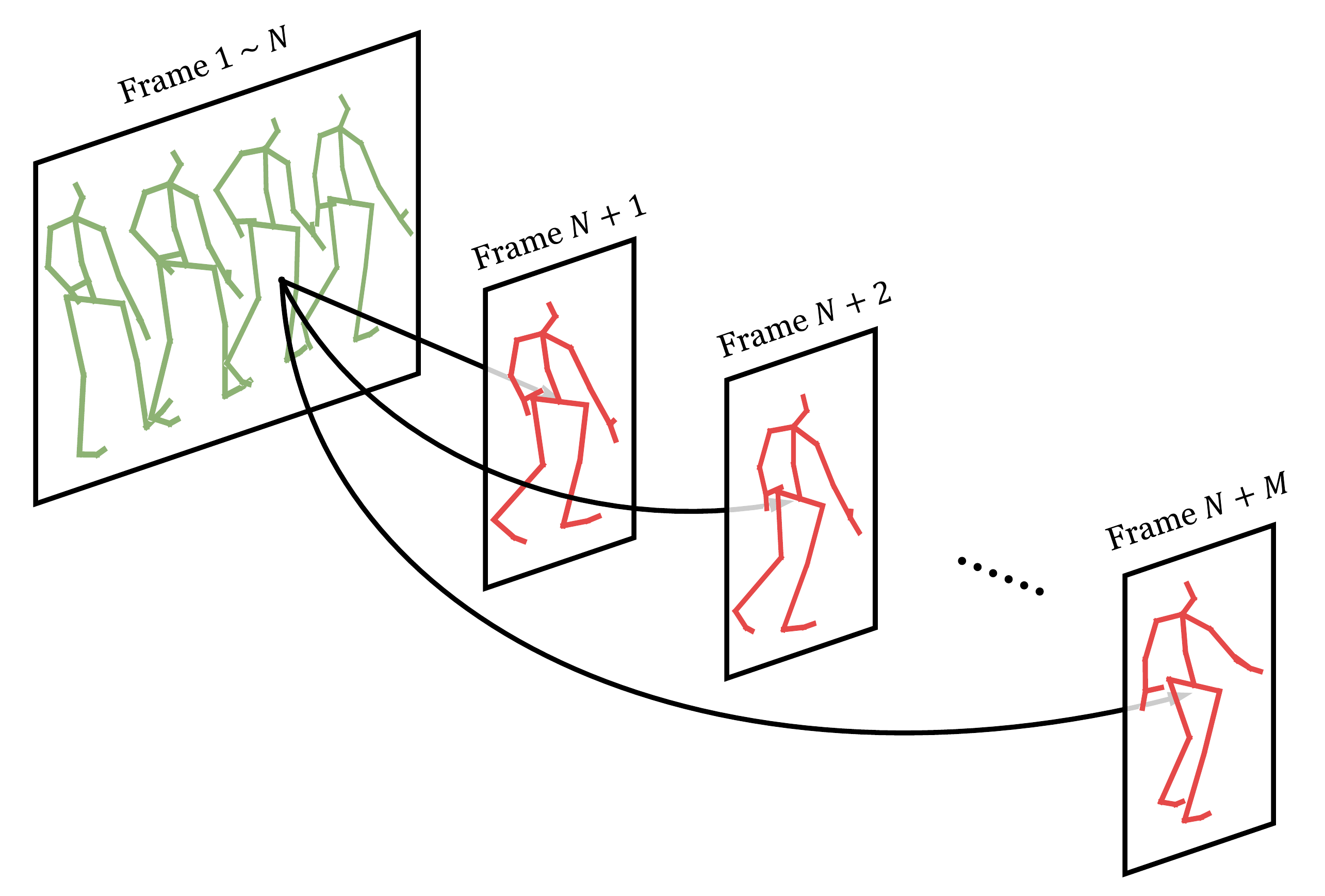}
    \caption{
      Given a real human motion sequence (in green color), we independently predict each frame of future data (in red color) in a complete non-autoregressive scheme, 
      in which the subsequent predicted frame would not be affected by the accuracy of the preceding one.}
  \label{fig:main}
 \end{figure}

  Therefore, the above autoregressive decoding pipeline for human motion prediction over the subsequent frames often suffers from the misguidance from the preceding prediction results and the evaluation criteria, the continuity and diversity of generated frames, are not guaranteed as well. 
  A natural question is whether we have a more feasible decoding pipeline, \emph{e.g.}, breaking the temporal dependency, that allows the subsequent frames to bypass the preceding ones directly during decoding.
  
\begin{figure*}[!t]
   \centering
     \includegraphics[width=0.93\linewidth]{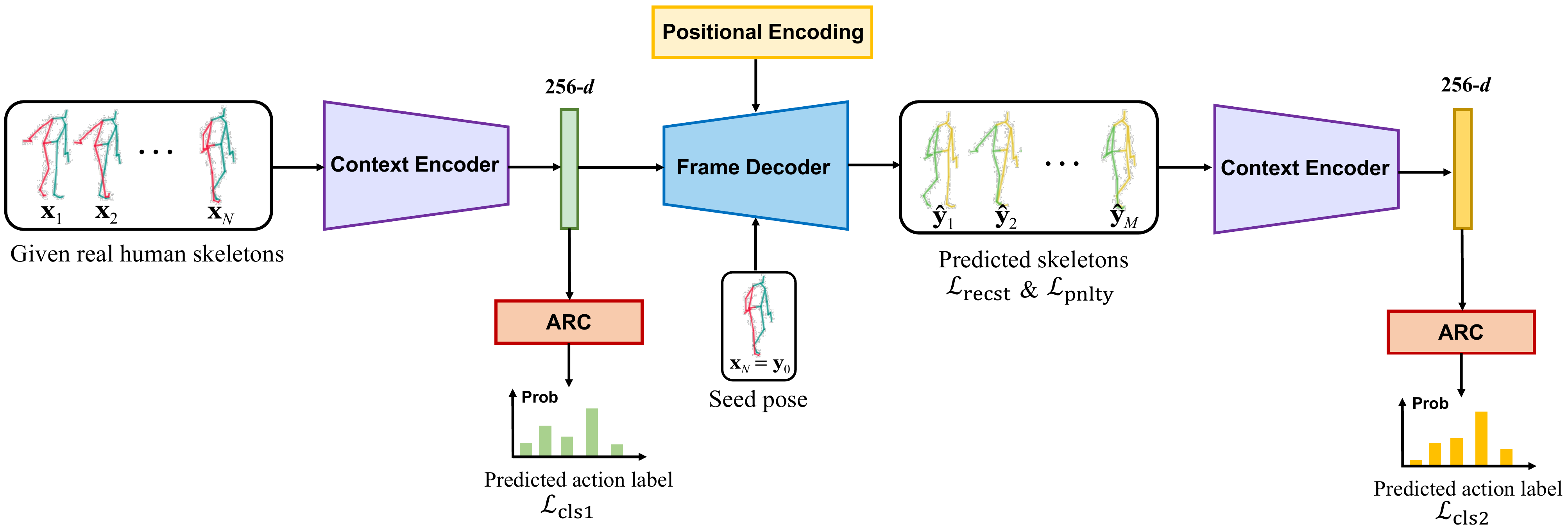}
     \caption{
       \textbf{Overview of Multitask Non-AuToregressive Model (mNAT).} 
       A real human skeleton sequence is first sent to context encoder, which is composed of multiple GCN-TCN blocks with residual connection, to obtain the 256-$d$ context feature (in green color).
       This feature is further sent the frame decoder as well as the seed pose and positional encoding vectors to generate each future frame independently.
       The frame decoder owns the same structure with the context encoder except that the kernel size is 1 in frame decoder while 9 in context encoder.
       Both the given real skeletons and predicted skeletons are sent to the same action recognition classifier (ARC) to predict the action category.
       Note that the two context encoders (in purple color) are the same one and so are the two classifiers (in red color).} 
   \label{fig:overview}
  \end{figure*}

  To this end, we propose a novel Non-AuToregressive framework (NAT), which largely eases the aforementioned issue.
  Specifically, NAT is composed of a context encoder (embedding the given poses), a frame decoder (predicting each future skeleton independently), and a positional encoding module (indicating temporal order).
  \emph{Context encoder} is modeled from both temporal and spatial perspectives by a TCN-based (Temporal Convolutional Network)\cite{oord2016wavenet} temporal encoder and a GCN-based (Graph Convolutional Network)\cite{kipf2017semi} spatial encoder through a skeleton kinematic tree, respectively.
  In principle, it encodes the existing real skeleton sequence data into a context feature space.
  In addition, \emph{frame decoder} sets up a prediction model to forecast each generated frame based on its corresponding direct connections to the existing real frames, as shown in Fig. \ref{fig:main}.
  Since temporal dependency is broken under our non-autoregressive setting, inspired by recent success from natural language processing\cite{vaswani2017attention}, we also propose a \emph{positional encoding module} which outputs a combination of sinusoidal waves with different frequencies as the representation of position.
  This representation can be viewed as a trajectory with physical constraints in the code space, and the frame decoder generates frames with the fusion of the representation and the context feature. The physical constraints, \emph{i.e.}, sinusoidal waves with different frequencies, guarantee the continuity and diversity. Therefore, the quality of the generated frames improves according to the evaluation criteria.
  Meanwhile, such a non-autoregressive setting enables parallel processing of multiple frames during decoding instead of sequential frame decoding used by conventional recurrent approaches.

  In addition, although achieving promising success, previous work rarely investigated the relation between the low-level human skeletons and the high-level human action category.
  The motivation behind this is that unlike deep models, human beings always make intention first and then perform it.
  If the model knows which action is to be generated, the forecasting would be much easier.
  For example, ``smoking" and ``phoning" may have very similar beginning. It is unlikely to forecast the following frames without knowing the exact action category.
  To this end, inspired by recent success on skeleton-based action recognition\cite{Yan_2019_STGCN, Shi1_2019_CVPR, Si_2019_CVPR, Shi2_2019_CVPR}, we propose a simple yet effective multitask training paradigm, namely Multitask Non-AuToregressive model (mNAT), which is further empowered by the merit of action recognition.
  Specifically, as shown in Fig. \ref{fig:overview}, we build a shared action recognition classifier (ARC) for both given real human skeletons and predicted ones, ensuring that our model is capable of predicting both the low-level and high-level future information.
  The experimental results show that the human motion prediction task achieves an obvious promotion for this multitasking scheme.

  In summary, the main contributions of this work are summarized as follows

  \begin{enumerate}
    \item We propose to solve the human motion prediction task with a novel Non-AuToregressive model (NAT), which largely alleviates the error accumulation problem.
    \item We further present a multitask training paradigm which is empowered by the merit of action recognition to predict both the low-level human skeletons and high-level human action category.
    \item Extensive experiments on both Human3.6M \cite{ionescu2013human3} and CMU-Mocap\footnote{\url{http://mocap.cs.cmu.edu}} benchmarks yield state-of-the-art results.
  \end{enumerate}

\section{Related Work}
\label{sec:relat}

\subsection{Human Motion Prediction}

Human motion prediction is a classical and challenging problem which has long been studied over years.
Previously, a series of statistical models have been applied to model action-agnostic human motion, including Hidden Markov Model (HMM)\cite{Lehrmann_2014_CVPR}, Gaussian Process Latent Variable Model (GPLVM)\cite{wang2007gaussian, yao2011learning}, Conditional Restricted Boltzmann Machine (CRBM)\cite{taylor2007modeling}, \emph{etc}.
However, these models suffered from difficulties such as high-dimensionality, complicated human motion nature, and approximate inference.

In recent years, deep neural networks have achieved success in every corner of computer vision \cite{sun2014deep, simonyan2014two, ren2015faster, chen2017deeplab}.
Holden \emph{et al}. \cite{holden2016deep} showed that human motion can be formulated as manifold via auto-encoders.
Due to the temporal nature of human motion prediction problem, much of current state-of-the-art work is based on RNN-based encoder-decoder structure.
Fragkiadaki \emph{et al.} \cite{Fragkiadaki_2015_ICCV} proposed Encoder-Recurrent-Decoder (ERD), which, for the first time, maps the pose into hidden state and propagates through an LSTM layer.
Also, Structural-RNN \cite{Jain_2016_CVPR} was presented to formulate human motion sequence as spatio-temporal graphs.
Martinez \emph{et al.} \cite{Martinez_2017_CVPR} designed a residual-based GRU (Res-GRU) model by predicting the relative residual between two consecutive frames instead of absolute skeleton.
This residual modeling works so well that it becomes the \emph{de facto} standard for the subsequent work.

Since Martinez's seminal work, human motion prediction has been roughly categorized into two groups.
The first group of approaches tend to seek better representations of human skeleton. 
This group of models explore either the spatial dependency or the different representation of each joint, \emph{e.g.}, exponential map, Euler angle.
Li \emph{et al.} \cite{Li_2018_CVPR} introduced a convolutional neural networks to model spatial and temporal dependency via a rectangle receptive field.
Guo \emph{et al.} \cite{Guo_2019_human} designed SkelNet which divides a human skeleton into five non-overlapping parts.
Similarly, Aksan \emph{et al.} \cite{Aksan_2019_ICCV} proposed an SP-layer which explicitly decomposes the pose into individual joints and could be further interfaced with a variety of baseline architectures.
Liu \emph{et al.} \cite{Liu_2019_CVPR} proposed to encode anatomical constraint explicitly with a Lie algebra representation.
Recently, Discrete Cosine Transform (DCT)\cite{Mao_2019_ICCV} was first introduced to encode temporal information via a series of DCT coefficients and achieved the state-of-the-art result so far.
Conventionally, human motion is represented as a skeletal kinematic tree, of which each joint is interpreted as a relative rotation on its parent joint.
To evaluate the effectiveness of different joint representation, QuaterNet \cite{Pavllo_2018_BMVC, pavllo2019modeling} was introduced to replace the commonly used exponential map representation by quaternion, which avoids common rotational problems such as non-uniqueness, discontinuity, and gimbal locks\cite{grassia1998practical}.

The second group of methods explore better measurements between ground truth and predicted skeletons. In general, $L1$ or $L2$ distance is commonly used for measuring distance. 
Gui \emph{et al.} \cite{Gui_2018_ECCV} presented Adversarial Geometry-Aware encoder-decoder (AGED), which replaced commonly used measurement by two loss: an adversarial loss to ensure the reality, and a geodesic loss to better model the motion.
Since then, Hernandez \emph{et al.}\cite{Hernandez_2019_ICCV} presented Spatio-Temporal Motion Inpainting (STMI-GAN) which formulates human motion forecasting as an image inpainting problem and further solved it with an improved GAN structure.

However, as mentioned, all the previous work adopted recurrent-based encoder-decoder structure, while a novel non-autoregressive decoding is proposed in this paper to predict each future frame directly, without the influence of the accuracy of previously predicted frames.

\subsection{Non-autoregressive Models}

RNN-based models, such as LSTM\cite{hochreiter1997long} or GRU\cite{chung2014empirical}, achieved great success in sequence modeling, especially in Neural Machine Translation (NMT)\cite{Cho_2014_properties, bahdanau2014neural, sutskever2014sequence}. 
In general, these methods generate tokens in a sequential manner, \emph{i.e.}, the new output word is dependent on the previously generated output.
Such left-to-right decoding manner suffers from problems like low efficiency and error accumulation \cite{bengio2015scheduled, lamb2016professor}.
To remedy these issues, several efforts have been paid on avoiding recurrence in sequence modeling.
Gehring \emph{et al.} \cite{gehring2017convolutional} proposed a sequence model based entirely on convolutional neural networks.
Vaswani \emph{et al.} \cite{vaswani2017attention} proposed the Transformer network, which stacks multiple self-attention layers to model the dependency on each token pairs.
Gu \emph{et al.} \cite{gu2017non} proposed non-autoregressive transformer that makes use of fertilities which represents how many times each source tokens are copied.
Recently, Yang \emph{et al.} \cite{yang2019non} introduced the non-autoregressive decoding into video captioning with an iterative refinement procedure during inference.

\begin{figure}[t]
   \centering
     \includegraphics[width=0.9\linewidth]{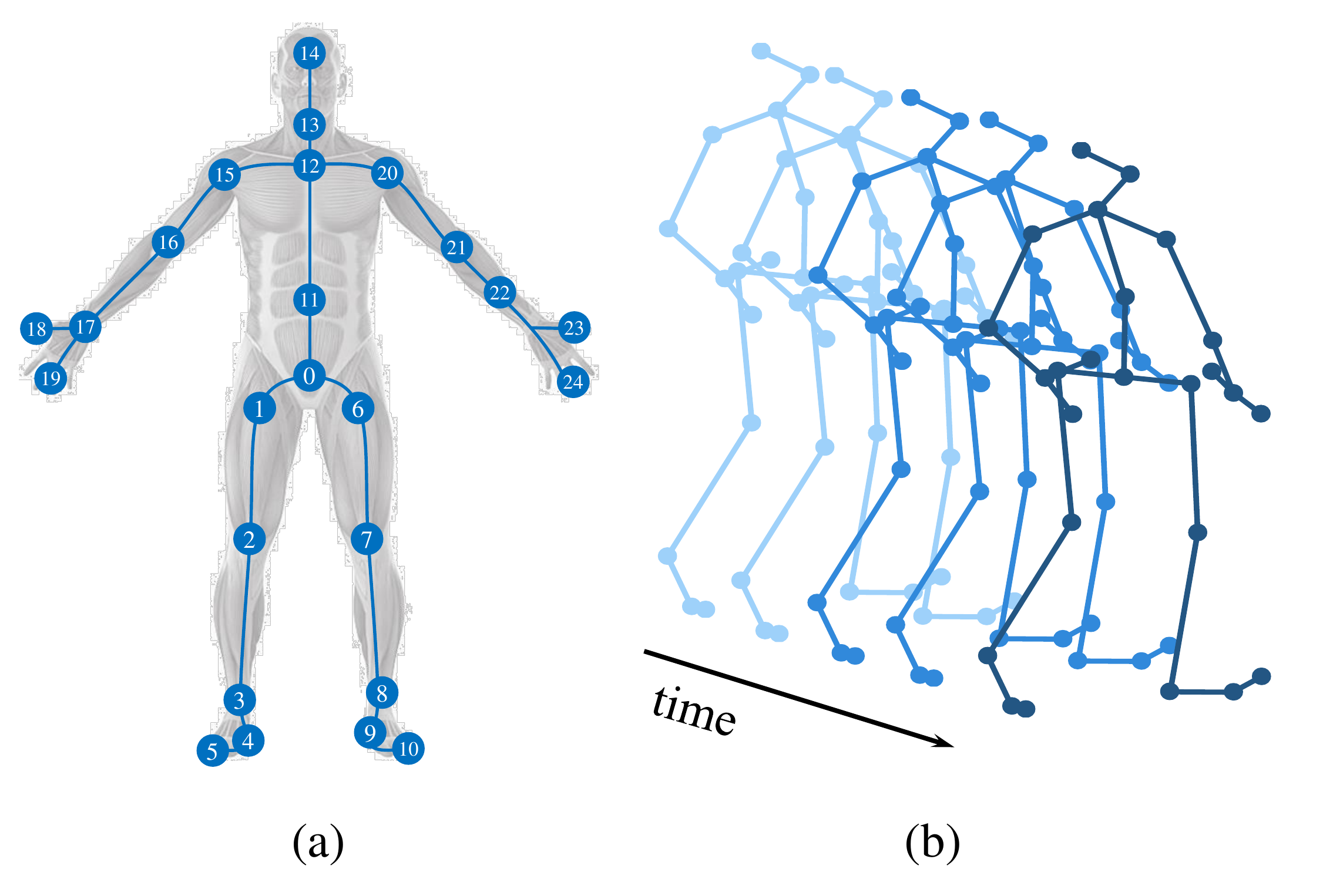}
     \caption{
       (a) Illustration of human skeleton in Human3.6M dataset. The blue circles indicate the detailed joint indexes.
       (b) Illustration of human motion sequence.
     }     
   \label{fig:skeleton}
\end{figure}

\begin{figure*}[!t]
   \centering
     \includegraphics[width=0.85\linewidth]{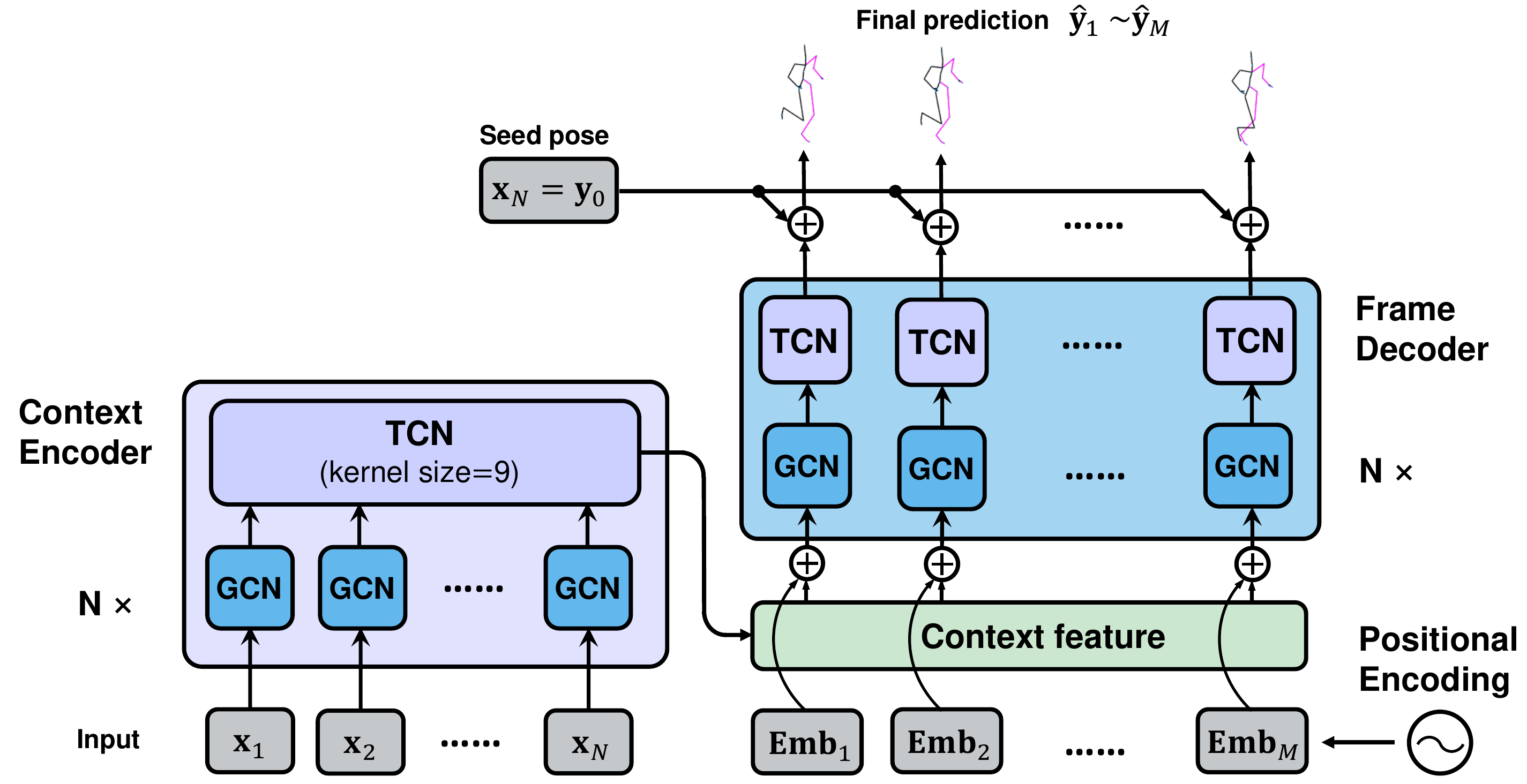}
     \caption{
       \textbf{Detail structure of NAT.} 
       Given the prefix human motion sequence $\mathbf{x}_{1:N}$, context encoder stacks GCN module and TCN module multiple times to encode it as context feature.
       The context feature is further added by a series of sinusoidal positional-related signal generated by positional encoding module.
       The features are then sent to frame decoder to generate final predicted human motion sequence $\hat{\mathbf{y}}_{1:M}$ in a non-autoregressive scheme.}
       
   \label{fig:detail}
\end{figure*}

As a sequence modeling task, human motion prediction shares similar nature with NMT and meanwhile possesses unique characteristics.
1) Human motion prediction is usually formulated as a continuous prediction task rather than a discrete one.
Hence, several useful tricks like Beam Search \cite{graves2012sequence, sutskever2014sequence} in NLP could not be directly adopted.
2) Human motion sequence contains rich skeletal structure information, which natural language token rarely owns.
Therefore, in this paper, we propose a GCN equipped context encoder to learn the rich skeletal structure as well as a frame decoder to implement non-autoregressive decoding.

\section{Problem Formulation}
\label{sec:background}


As shown in Fig. \ref{fig:skeleton} (a), the human motion skeleton is usually represented as a skeletal kinematic tree.
A kinematic tree is composed of one root joint and several other joints as child nodes.
Each child node possesses only one parent joint, forming a tree structure.
Hence, the human motion sequence is constructed by stacking multiple human skeletons through time horizon, as shown in Fig. \ref{fig:skeleton} (b).

Specifically, we consider to be given a length-$N$ observed sequence $\mathbf{X} = (\mathbf{x}_1, \mathbf{x}_2, ..., \mathbf{x}_N) \in \mathbb{R}^{N\times J\times K}$, where each of the frames $\mathbf{x}_n = \{\mathbf{x}_n^j\}_{j=1}^{J}$ represents the single skeleton, containing $J$ joints data. 
$\mathbf{x}_n^j \in \mathbb{R}^K$ is a minimal per-joint representation at the $n$-th frame and the $j$-th joint, of which $K$ is the feature dimension which represents human joint data.
In this paper, we adopt $K=4$ for quaternion as this format is free of discontinuity and singularity\cite{Pavllo_2018_BMVC}.
Our goal is to predict consecutive length-$M$ target sequence $\mathbf{Y} = (\mathbf{y}_1, \mathbf{y}_2, ..., \mathbf{y}_M)$. Note that decoding always starts from last frame of the given  sequence $\mathbf{X}$. For simplicity, we name it as ``seed pose":  $\mathbf{x}_{\mathrm{seed}}=\mathbf{x}_N=\mathbf{y}_0$. 

In the following, we first explain the reason why autoregressive model leads to error accumulation problem.
We then discuss each part of our NAT model in detail.
In particular, we introduce context encoder, positional encoding, and frame decoder in Sec \ref {sec:context}, \ref{sec:positional}, \ref{sec:frame}, respectively.
In Sec \ref{sec:optimization}, we introduce the multitask training pipeline for our NAT model.

Usually, the human motion is typically viewed as an inertial model, where only small changes happen in two consecutive frames.
Therefore, RNNs are adopted to model this temporal continuity in an autoregressive paradigm.
In this paper, we argue that this inertance still exists in short term (less than one second).
Hence, instead of predicting the target skeleton in a frame-by-frame manner, we directly regress the residual item between each target pose and seed pose.

In detail, assume the conventional RNN-based  encoder-decoder framework acts as


   

\begin{equation}
   \hat{\mathbf{y}}_{t} = 
   \hat{\mathbf{y}}_{t-1} +
   \mathbb{D}\big(
   \hat{\mathbf{y}}_{1:t-1}, 
   \mathbb{E}(\mathbf{X})
   \big),
   \label{eqn:reccurent}
\end{equation}

\noindent where $\mathbb{E}(\cdot)$ denotes the encoder function, $\mathbb{D}(\cdot)$ is the decoder function, and $\hat{\mathbf{y}}_t$ is the predicted pose at time $t$. 

Note that the above equation defines the recursive formula between two consecutive frames. We then expand this equation by

\begin{equation}
   \begin{aligned}
   \hat{\mathbf{y}}_{t} &= 
   \hat{\mathbf{y}}_{t-1} +
   \mathbb{D}\big(
   \hat{\mathbf{y}}_{1:t-1}, 
   \mathbb{E}(\mathbf{X})
   \big) \\
   &= \hat{\mathbf{y}}_{t-2} +
   \mathbb{D}\big(
   \hat{\mathbf{y}}_{1:t-2}, 
   \mathbb{E}(\mathbf{X})
   \big) +
   \mathbb{D}\big(
   \hat{\mathbf{y}}_{1:t-1}, 
   \mathbb{E}(\mathbf{X})
   \big) \\
   &= \mathbf{y}_0 +
   \sum_{i=0}^{t-1} 
   \mathbb{D}\big(
   \hat{\mathbf{y}}_{1:i}, 
   \mathbb{E}(\mathbf{X})
   \big),
   \end{aligned}
   \label{eq:3}
\end{equation}

\noindent where $\mathbf{y}_0$ is the seed pose, which is the last frame of the given motion sequence as we discussed above. 
For consistence of equation form, we assume $\hat{\mathbf{y}}_{1:0} = \mathbf{y}_0$ when we calculate $\hat{\mathbf{y}}_{1} = \mathbf{y}_0 + \mathbb{D}\big(\mathbf{y}_0, \mathbb{E}(\mathbf{X})\big)$.

We argue that the aforementioned error accumulation problem could be derived from Eq. \ref{eq:3}. 
First of all, Eq. \ref{eq:3} constructs the relation between the target pose $\hat{\mathbf{y}}_{t}$ and seed pose $\mathbf{y}_{0}$. 
The residual item between $\hat{\mathbf{y}}_{t}$ and $\mathbf{y}_{0}$ is a sum of multiple predictions.
Suppose an initial error item $\delta_1$ happens between $\hat{\mathbf{y}}_{1}$ and $\mathbf{y}_{1}$.
Then we have $\hat{\mathbf{y}}_{2} = \mathbf{y}_{1} + \delta_1 + \mathbb{D}\big(\mathbf{y}_{1} + \delta_1, \mathbb{E}(\mathbf{X})\big)$, which means the initial error term $\delta_1$ propagates to $\delta_2$.
Therefore, the initial error term $\delta_1$ would spread to each time item up to $\delta_t$ repeatedly.
Hence, the error accumulation happens when the sum item increases rapidly with the evolution of the time $t$.
This per-frame prediction error grows so fast that the long-term prediction soon becomes implausible to use.

Inspired by this observation, we model the residual term between $\hat{\mathbf{y}}_{t}$ and $\mathbf{y}_{0}$ directly, rather than modeling it in an autoregressive way. 
Specifically, $\hat{\mathbf{y}}_{t}$ no longer depends on the accuracy of previous generated poses in our design. 
We directly obtain each target pose based on the last available ground truth skeleton $\mathbf{y}_{0}$ only, 
which largely alleviates the error accumulation problem.


\section{Non-autoregressive Model}
\label{sec:nat}

In this section, we provide the details of our NAT structure. 
Unlike complicatedly designed models in natural language processing, we find that non-autoregressive architecture could be simply implemented by three components: a context encoder $\mathbb{E}(\cdot)$, a frame decoder $\mathbb{D}(\cdot)$, and a positional encoding module, as shown in Fig. \ref{fig:detail}.
We introduce each part in detail below.

\subsection{Context Encoder}
\label{sec:context}
%
Usually, a human motion sequence is represented as a spatio-temporal graph. 
Therefore, the encoder needs to simultaneously model both the joint-wise dependency in the spatial domain and frame-wise dependency in the temporal domain.   
To this end, we propose to stack multiple GCN-TCN blocks to form the context encoder which is capable of generating context feature being representative of the whole given sequence.

To encode the spatial dependency of human skeletons, we make use of GCNs\cite{kipf2017semi}. 
GCNs are a class of models which are specially designed for non-Euclidean data. 
To make our paper self contained, we briefly introduce how GCNs work here.
As mentioned, each frame of human skeleton sequence contains $J$ joints. 
The bone connection is thus formulated as an adjacency matrix $\mathbf{A} \in \mathbb{R}^{J \times J}$, where $\mathbf{A}_{ij} = 1$ if and only if joint $i$ connects with joint $j$ (each joint connects with itself). 
Assume that in layer $l$, we have input feature as $\mathbf{h}^{(l)} \in \mathbb{R}^{J \times K}$, where $K$ denotes input dimension.
Following Kipf \emph{et al.}\cite{kipf2017semi}, we adopt first order approximation and output $\mathbf{h}^{(l+1)}$ as

\begin{equation}
   \mathbf{h}^{(l+1)} = \sigma
   \Big(
      \mathrm{BN}\big(
      \tilde{\mathbf{A}}
      \cdot \mathbf{h}^{(l)}
      \cdot \mathbf{W}^{(l)}
      \big)   
   \Big),
\end{equation} 

\noindent where $\tilde{\mathbf{A}} = \mathbf{D}^{-\frac{1}{2}}\mathbf{A}\mathbf{D}^{-\frac{1}{2}}$ is the normalized adjacency matrix 
and $\mathbf{D}_{ii} = \sum_{j} \mathbf{A}_{ij}$ is the corresponding degree matrix, 
$\mathbf{W}^{(l)} \in \mathbb{R}^{K\times K}$ is a trainable weight matrix, 
$\textrm{BN}(\cdot)$ denotes Batch Normalization\cite{ioffe2015batch},
and $\sigma(\cdot)$ is the Leaky ReLU\cite{maas2013rectifier}.

To encode the temporal dependency, we make use of TCN\cite{oord2016wavenet} where RNNs are replaced by 1D CNNs.
To capture the long-term dependency, multiple CNN layers are stacked to increase the receptive field.
In the TCN model, the receptive field drastically grows in a linear speed.
%
%
%
%
%
In practice, we find a large kernel size is necessary to ensure that the final feature covers the whole input sequence.
However, once reaching an appropriate receptive field, no benefit is found by using a larger kernel size.
We conduct experiments in Sec \ref{sec:exper} to verify the effects of different kernel sizes of TCN.

\begin{figure}[!t]
   \centering
     \includegraphics[width=0.85\linewidth]{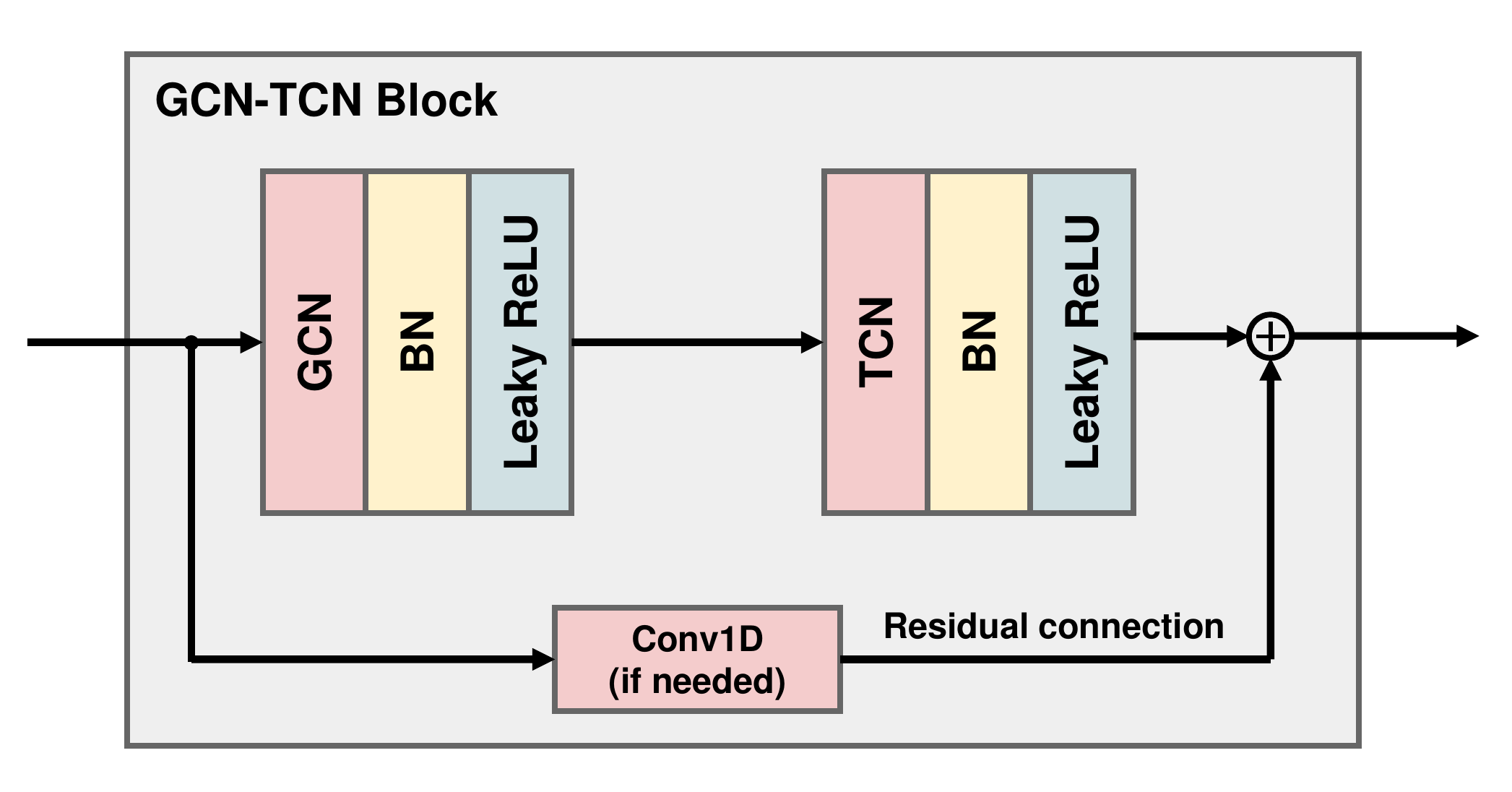}
     \caption{
       Illustration of the GCN-TCN Block. Both GCN and TCN operation are followed by a BN and Leaky ReLU. Note that the below Conv1D only appears when input channel is not equal to output channel.
       }
       
   \label{fig:GCN-TCN}
  \end{figure}

We also add skip-connection\cite{He_2016_CVPR} between each two blocks as it makes easier to propagate the gradients and accelerates training, as shown in Fig. \ref{fig:GCN-TCN}.
The number of channels is 64, 64, 128, 128, 256, 256 in total 6 blocks, respectively, which map the input 4-$d$ quaternion to a 256-$d$ feature.
In the end, we perform global average pooling in both temporal dimension and spatial dimension to obtain a single context feature $\mathbf{c} = \mathbb{E}(\mathbf{X}) \in \mathbb{R}^{256}$.

\subsection{Positional Encoding Module}
\label{sec:positional}
Compared with the autoregressive models, which implicitly encode temporal order in a frame-by-frame way, the non-autoregressive model faces a problem on representing time.
To this end, we propose to inject the explicit temporal signal into decoder directly.
We further rewrite Eq. \ref{eqn:reccurent} as follows
\begin{equation}
   \hat{\mathbf{y}}_{t} = \mathbf{y}_{0} +
   \mathbb{D}\big(\mathbf{p}(t), \mathbb{E}(\mathbf{X})\big),
   \label{eqn:5}
\end{equation}

\noindent where $t$ represents the time index and $\mathbf{p}(\cdot)$ is a function that maps input scalar index into a vector form embedding. In our case, we call this mapping function $\mathbf{p}(\cdot)$ as positional encoding module.

Following the former success of Transformer \cite{vaswani2017attention}, we adopt the sinusoidal functions of different frequencies as our positional encoding.
In particular, we aim to map relative time index into a vector form feature.
For time index $t=1,...,M$, the positional encoding function is expressed as

\begin{equation}
   \begin{aligned}
      & p_{2i}(t) = \sin\big(\alpha \cdot t/\beta^{2i/d_{\mathrm{model}}}\big) \\
      & p_{2i+1}(t) = \cos\big(\alpha \cdot t/\beta^{2i/d_{\mathrm{model}}}\big),
   \end{aligned}
   \label{eqn:pos_enc_modify}
\end{equation}

\noindent where $p_{2i}(t)$, $p_{2i+1}(t)$ represent the even and odd dimension of $\mathbf{p}(t)$, $d_{\mathrm{model}}$ denotes the positional embedding dimension, $\alpha$ is a scale factor which controls the difference across time indexes, $\beta$ controls the wavelength for each dimension, and $i$ is the dimension index ranging from 1 to $\lfloor  d_{\mathrm{model}} / 2 \rfloor$.

The benefit of sinusoidal positional encoding module is two-fold. 
Firstly, compared with trivial one-hot encoding, this encoding outputs a continuous form vector, carrying more information.
Secondly, the sinusoidal positional embedding is highly correlated, \emph{i.e.}, the closer two time indexes $t_1$, $t_2$ are, the more similar $\mathbf{p}(t_1)$, $\mathbf{p}(t_2)$ are.
Consequently, the positional embedding could be seen as a disturbance item added to the encoded feature, ensuring the smoothness of generated sequence.

\begin{figure}[!t]
   \centering
     \includegraphics[width=0.8\linewidth]{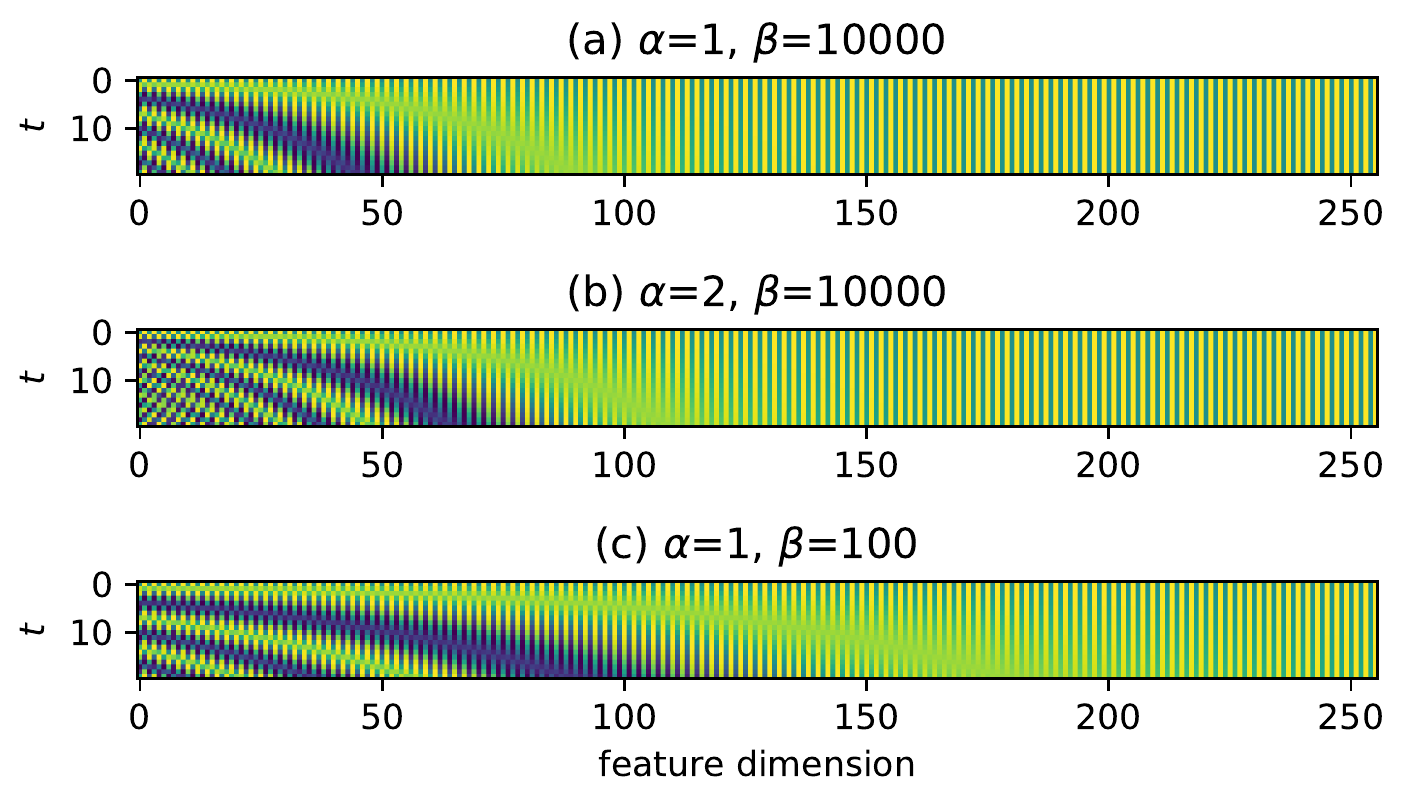}
     \caption{
       Illustration of the Positional Embedding. This embedding is generated by sine and cosine functions of different frequencies. We also explore the influence of hyper-parameter $\alpha$ and $\beta$.
       }
   \label{fig:pos_emb}
\end{figure}

Note that in the original paper, the $\alpha$ and $\beta$ are set to fixed values ($\alpha=1$, $\beta=10000$).
We argue that due to the domain difference, \emph{i.e.}, the predicted length is usually 10 or 25 in human motion prediction while hundreds of tokens might be involved in neural machine translation, the default setting could be sub-optimal in this task.
As shown in Fig. \ref{fig:pos_emb}, with the growth of $\alpha$, the difference between each embedding increases (see (a), (b)). 
With the decrease of $\beta$, more dimensions are involved to distinguish different embeddings.
We conduct extensive experiments on exploring the influence of $\alpha$ and $\beta$.

\subsection{Frame Decoder}
\label{sec:frame}
As mentioned above, the frame decoder is responsible for generating each frame independently.
To generate $M$ future frames, we first obtain a series of positional embeddings $\mathbf{P} = \{\mathbf{p}(1), ..., \mathbf{p}(M)\}$, each of which is also a 256-$d$ vector.
These embeddings are further added to the context feature $\mathbf{c}$ to form the input of the frame decoder $\mathbf{F} = \{\mathbf{f}_t\}_{t=1}^{M}$, where each $\mathbf{f}_t = \mathbf{c} + \mathbf{p}(t)$.

For simplicity, we reuse the GCN-TCN blocks as in context encoder.
The only difference lies in that the temporal kernel size is set to 1 in each TCN operation.
Under this setting, we rewrite Eq. \ref{eqn:5} as: $\hat{\mathbf{y}}_{t} = \mathbf{y}_{0} + \mathbb{D}(\mathbf{f}_t)$,
where each frame generation process is strictly limited to the single frame, avoiding to be affected by subsequent predicted frames, also shown in Fig. \ref{fig:detail}.
Similar to context encoder, the number of channels is 256, 128, 128, 64, 64, 4 with total 6 blocks, respectively, which map the input 256-$d$ feature back to 4-$d$ quaternion.

\section{Multitask Training}
\label{sec:optimization}

\subsection{Action Recognition Classifier}
As another classic task in skeleton-based activity understanding, skeleton-based action recognition also attracts lots of attention recently.
However, few work has explored the relation of these two tasks. 
Martinez \emph{et al.}\cite{Martinez_2017_CVPR} proposed Res-GRU MA (Multi-Action) by simply concatenating one-hot vectors with 15 action classes of Human3.6M dataset. 
The result shows limited performance gain.
Therefore, most of the subsequent work reports the SA (Single-Action) result without the action category information.

In this paper, we argue that, as human beings, the high-level human action category guides the low-level human skeletons and the existing literatures rarely investigate these two related tasks. 
To this end, we propose an action recognition classifier (ARC) on the top of the context feature $\mathbf{c}$, as shown in Fig. \ref{fig:overview}.
The ARC module is implemented with a three-layer MLP, where FC, Dropout, and LeakyReLU are included. Let the ground truth label be $\mathbf{y}_{\mathrm{cls}}=\{0, 1\}^{C}$, where $C$ denotes the number of actions in total, and the corresponding classification result be $\mathbf{o}_1 = \textrm{Softmax}(\textrm{ARC}(\mathbf{c}))$. The classification loss is formulated as

\begin{equation}
   \mathcal{L}_{\mathrm{cls1}} =
   -\mathbf{y}_{\mathrm{cls}}^\top
   \log(\mathbf{o}_1).
   \label{eqn:cls1}
\end{equation}

Inspired by self-supervised learning \cite{Doersch_2015_ICCV, Gomez_2017_CVPR}, we also add a cycle consistency classification loss. 
We expect that the predicted human motion sequence $\hat{\mathbf{Y}} = \{\hat{\mathbf{y}}_t\}_{t=1}^{M}$ not only is close to the ground truth sequence, but also represents the high-level action category.
Therefore,  $\hat{\mathbf{Y}}$ is sent to the same context encoder and ARC to obtain the classification result $\mathbf{o}_2$. The cycle consistency classification loss is
\begin{equation}
   \mathcal{L}_{\mathrm{cls2}} =
   -\mathbf{y}_{\mathrm{cls}}^\top
   \log(\mathbf{o}_2).
   \label{eqn:cls2}
\end{equation}

\begin{algorithm}[t]
   \caption{NAT Multitask Training}
   \label{alg1}
   \begin{algorithmic}[1]
      \renewcommand{\algorithmicrequire}{\textbf{Input:}}
      \renewcommand{\algorithmicensure}{\textbf{Output:}}
      \REQUIRE Training set $\mathcal{D}=\{(\mathbf{X}^i, \mathbf{Y}^i, \mathbf{y}_{\mathrm{cls}}^i)\}_{i=1}^{\#sample}$ with $C$ classes; Training iterations $\tau$.
      \ENSURE The parameters $\theta$ of NAT and ARC
      \FOR{iteration = $1, ..., \tau$}
         \STATE Randomly sample $(\mathbf{X}^i, \mathbf{Y}^i, \mathbf{y}_{\mathrm{cls}}^i)$ from $\mathcal{D}$
         \STATE Compute context feature $\mathbf{c}$ with context encoder
         \STATE Obtain the predicted sequence $\hat{\mathbf{Y}}^i$ using Eq. \ref{eqn:5}
         \STATE Compute $\mathbf{o}_1$ from $\mathbf{c}$ with ARC module
         \STATE Compute $\mathbf{c}'$ by feeding $\hat{\mathbf{Y}}^i$ back into context encoder
         \STATE Compute $\mathbf{o}_2$ from $\mathbf{c}'$ with ARC module
         \STATE Calculate $\mathcal{L}_{\mathrm{recst}}$, $\mathcal{L}_{\mathrm{pnlty}}$, $\mathcal{L}_{\mathrm{cls1}}$, and $\mathcal{L}_{\mathrm{cls2}}$ using Eq. \ref{eqn:cls1} $\sim$ \ref{eqn:pnlty}
         \STATE Update the parameters $\theta$ using Eq. \ref{eqn:loss_all} by ADAM\cite{kingma2014adam}

      \ENDFOR
      \RETURN The parameters $\theta$ of NAT and ARC

   \end{algorithmic}
\end{algorithm}

\begin{table}[]
   \begin{center}
   \caption{Comparison of mean joint error of angle space on average of all 15 actions of Human3.6M dataset}
   \label{tab:avg}
      \begin{tabular}{l|llll|c}
         \Xhline{1pt}
                                                & \multicolumn{4}{c|}{\textbf{Average}}                                                                 & \multirow{2}{*}{\textbf{Conference}} \\
         \textbf{millseconds}                   & \multicolumn{1}{c}{80} & \multicolumn{1}{c}{160} & \multicolumn{1}{c}{320} & \multicolumn{1}{c|}{400} & \textbf{}           \\ \Xhline{1pt}
         Zero-velocity\cite{Martinez_2017_CVPR} & 0.42                   & 0.74                    & 1.12                    & 1.20                     & CVPR2017            \\ \hline
         Res-GRU\cite{Martinez_2017_CVPR}       & 0.39                   & 0.72                    & 1.08                    & 1.22                     & CVPR2017            \\
         ConvSeq2Seq\cite{Li_2018_CVPR}         & 0.38                   & 0.68                    & 1.01                    & 1.13                     & CVPR2018            \\
         QuaterNet\cite{Pavllo_2018_BMVC}       & 0.35                   & 0.64                    & 1.07                    & 1.23                     & BMVC2018            \\
         AGED w/ adv\cite{Gui_2018_ECCV}         & 0.33                   & 0.58                    & 0.94                    & 1.01                     & ECCV2018            \\
         SkelNet\cite{Guo_2019_human}           & 0.36                   & 0.64                    & 0.99                    & 1.02                     & AAAI2019            \\
         TD-DCT\cite{Mao_2019_ICCV}             & \textbf{0.27}          & 0.51                    & 0.83                    & 0.95                     & ICCV2019            \\ \hline
         NAT (Ours)                           & \textbf{0.27}          & 0.50                    & 0.79                    & 0.91                     & -                   \\
         mNAT (Ours)                             & \textbf{0.27}          & \textbf{0.48}           & \textbf{0.74}           & \textbf{0.85}            & -                   \\ \Xhline{1pt}
      \end{tabular}
   \end{center} 
\end{table}

\subsection{Training}
We now summarize the whole training process.
Given the predicted human motion sequence $\hat{\mathbf{Y}} = \{\hat{\mathbf{y}}_t\}_{t=1}^{M}$ and ground truth, we apply the average $L1$ distance as the reconstruction loss

\begin{equation}
   \mathcal{L}_{\mathrm{recst}} = \frac{1}{J \times M}
   \sum_{j=1}^{J}
   \sum_{t=1}^{M}
   \big| \hat{\mathbf{y}}_t^j - \mathbf{y}_t^j \big|, 
   \label{eqn:recst}
\end{equation}

\noindent where $\hat{\mathbf{y}}_t^j$ denotes the predicted skeleton of the $j$-th joint in the $t$-th frame, and $\mathbf{y}_t^j$ is the corresponding ground truth.

Since that we use quaternion as the joint representation, we must ensure the output $\hat{\mathbf{y}}_t$ has unit length as only unit quaternion represents a valid 3D rotation \cite{pervin1982quaternions, shoemake1985animating}.
To this end, we also add a penalty loss for each of the prediction to ensure this property

\begin{equation}
   \mathcal{L}_{\mathrm{pnlty}} = \frac{1}{J \times M}
   \sum_{j=1}^{J}
   \sum_{t=1}^{M}
   \big(\lVert \hat{\mathbf{y}}_t^j \rVert_2^2-1\big)^2.
   \label{eqn:pnlty}
\end{equation}

To summarize, our overall objective is
\begin{equation}
   \mathcal{L} =
   \mathcal{L}_{\mathrm{recst}} +
   \lambda_{\mathrm{pnlty}}
   \mathcal{L}_{\mathrm{pnlty}} +
   \lambda_{\mathrm{cls}}
   (\mathcal{L}_{\mathrm{cls1}} + 
   \mathcal{L}_{\mathrm{cls2}}),
   \label{eqn:loss_all}
\end{equation}

\noindent where $\lambda_{\mathrm{pnlty}}$ and $\lambda_{\mathrm{cls}}$ control the relative importance of each loss item. The pseudo code of NAT Multitask Training is shown in Algorithm \ref{alg1}.

\begin{table*}[]
   \begin{center}
   \caption{Comparison of Mean Joint Error of angle space between our model and the state-of-the-art methods on all 15 actions of Human3.6M dataset}
   \label{tab:all}
      \resizebox{\textwidth}{!}{
         \begin{tabular}{lcccccccccccccccccccc}
            \Xhline{1pt}
            \multicolumn{1}{l|}{}                                       & \multicolumn{4}{c|}{\textbf{Walking}}                                                                   & \multicolumn{4}{c|}{\textbf{Eating}}                                                                    & \multicolumn{4}{c|}{\textbf{Smoking}}                                                                   & \multicolumn{4}{c|}{\textbf{Discussion}}                                                                & \multicolumn{4}{c}{\textbf{Direction}}                                                    \\
            \multicolumn{1}{l|}{\textbf{milliseconds}}                  & 80                   & 160                  & 320                  & \multicolumn{1}{c|}{400}           & 80                   & 160                  & 320                  & \multicolumn{1}{c|}{400}           & 80                   & 160                  & 320                  & \multicolumn{1}{c|}{400}           & 80                   & 160                  & 320                  & \multicolumn{1}{c|}{400}           & 80                   & 160                  & 320                  & 400                  \\ \Xhline{1pt}
            \multicolumn{1}{l|}{Zero-velocity\cite{Martinez_2017_CVPR}} & 0.39                 & 0.68                 & 0.99                 & \multicolumn{1}{c|}{1.15}          & 0.27                 & 0.48                 & 0.73                 & \multicolumn{1}{c|}{0.86}          & 0.26                 & 0.48                 & 0.97                 & \multicolumn{1}{c|}{0.95}          & 0.31                 & 0.67                 & 0.94                 & \multicolumn{1}{c|}{1.04}          & 0.39                 & 0.59                 & 0.79                 & 0.89                 \\
            \multicolumn{1}{l|}{Res-GRU\cite{Martinez_2017_CVPR}}       & 0.27                 & 0.47                 & 0.70                 & \multicolumn{1}{c|}{0.78}          & 0.25                 & 0.43                 & 0.71                 & \multicolumn{1}{c|}{0.87}          & 0.33                 & 0.61                 & 1.04                 & \multicolumn{1}{c|}{1.19}          & 0.31                 & 0.69                 & 1.03                 & \multicolumn{1}{c|}{1.12}          & 0.26                 & 0.47                 & 0.72                 & 0.84                 \\
            \multicolumn{1}{l|}{ConvSeq2Seq\cite{Li_2018_CVPR}}         & 0.33                 & 0.54                 & 0.68                 & \multicolumn{1}{c|}{0.73}          & 0.22                 & 0.36                 & 0.58                 & \multicolumn{1}{c|}{0.71}          & 0.26                 & 0.49                 & 0.96                 & \multicolumn{1}{c|}{0.92}          & 0.32                 & 0.67                 & 0.94                 & \multicolumn{1}{c|}{1.01}          & 0.39                 & 0.60                 & 0.80                 & 0.91                 \\
            \multicolumn{1}{l|}{AGED w/ adv\cite{Gui_2018_ECCV}}         & 0.22                 & 0.36                 & 0.55                 & \multicolumn{1}{c|}{0.67}          & 0.17                 & 0.28                 & 0.51                 & \multicolumn{1}{c|}{0.64}          & 0.27                 & 0.43                 & 0.82                 & \multicolumn{1}{c|}{0.84}          & 0.27                 & 0.56                 & 0.76                 & \multicolumn{1}{c|}{0.83}          & \textbf{0.23}        & \textbf{0.39}        & 0.63                 & 0.69                 \\
            \multicolumn{1}{l|}{SkelNet\cite{Guo_2019_human}}           & 0.31                 & 0.50                 & 0.69                 & \multicolumn{1}{c|}{0.76}          & 0.20                 & 0.31                 & 0.53                 & \multicolumn{1}{c|}{0.69}          & 0.25                 & 0.50                 & 0.93                 & \multicolumn{1}{c|}{0.89}          & 0.30                 & 0.64                 & 0.89                 & \multicolumn{1}{c|}{0.98}          & 0.36                 & 0.58                 & 0.77                 & 0.86                 \\
            \multicolumn{1}{l|}{TD-DCT\cite{Mao_2019_ICCV}}             & 0.18                 & 0.31                 & 0.49                 & \multicolumn{1}{c|}{0.56}          & \textbf{0.16}        & 0.29                 & 0.50                 & \multicolumn{1}{c|}{0.62}          & \textbf{0.22}        & 0.41                 & 0.86                 & \multicolumn{1}{c|}{0.80}          & \textbf{0.20}        & \textbf{0.51}        & 0.77                 & \multicolumn{1}{c|}{0.85}          & 0.26                 & 0.45                 & 0.71                 & 0.79                 \\ \hline
            \multicolumn{1}{l|}{NAT (Ours)}                           & 0.19                 & \textbf{0.28}        & \textbf{0.45}        & \multicolumn{1}{c|}{\textbf{0.51}} & \textbf{0.16}        & \textbf{0.25}        & \textbf{0.44}        & \multicolumn{1}{c|}{0.58}          & 0.23                 & 0.42                 & 0.82                 & \multicolumn{1}{c|}{0.84}          & 0.22                 & 0.54                 & 0.79                 & \multicolumn{1}{c|}{0.89}          & 0.26                 & 0.42                 & 0.62                 & 0.72                 \\
            \multicolumn{1}{l|}{mNAT (Ours)}                             & \textbf{0.17}        & 0.29                 & \textbf{0.45}        & \multicolumn{1}{c|}{0.53}          & 0.17                 & 0.31                 & 0.48                 & \multicolumn{1}{c|}{\textbf{0.54}} & \textbf{0.22}        & \textbf{0.40}        & \textbf{0.81}        & \multicolumn{1}{c|}{\textbf{0.78}} & 0.23                 & 0.54                 & \textbf{0.72}        & \multicolumn{1}{c|}{\textbf{0.80}} & 0.27                 & 0.43                 & \textbf{0.58}        & \textbf{0.67}        \\ \Xhline{1pt}
            \multicolumn{1}{l|}{}                                       & \multicolumn{4}{c|}{\textbf{Greeting}}                                                                  & \multicolumn{4}{c|}{\textbf{Phoning}}                                                                   & \multicolumn{4}{c|}{\textbf{Posing}}                                                                    & \multicolumn{4}{c|}{\textbf{Purchases}}                                                                 & \multicolumn{4}{c}{\textbf{Sitting}}                                                      \\
            \multicolumn{1}{l|}{\textbf{milliseconds}}                  & 80                   & 160                  & 320                  & \multicolumn{1}{c|}{400}           & 80                   & 160                  & 320                  & \multicolumn{1}{c|}{400}           & 80                   & 160                  & 320                  & \multicolumn{1}{c|}{400}           & 80                   & 160                  & 320                  & \multicolumn{1}{c|}{400}           & 80                   & 160                  & 320                  & 400                  \\ \Xhline{1pt}
            \multicolumn{1}{l|}{Zero-velocity\cite{Martinez_2017_CVPR}} & 0.54                 & 0.89                 & 1.30                 & \multicolumn{1}{c|}{1.49}          & 0.64                 & 1.21                 & 1.65                 & \multicolumn{1}{c|}{1.83}          & 0.28                 & 0.57                 & 1.13                 & \multicolumn{1}{c|}{1.37}          & 0.62                 & 0.88                 & 1.19                 & \multicolumn{1}{c|}{1.27}          & 0.40                 & 1.63                 & 1.02                 & 1.18                 \\
            \multicolumn{1}{l|}{Res-GRU\cite{Martinez_2017_CVPR}}       & 0.75                 & 1.17                 & 1.74                 & \multicolumn{1}{c|}{1.83}          & 0.23                 & 0.43                 & 0.69                 & \multicolumn{1}{c|}{0.82}          & 0.36                 & 0.71                 & 1.22                 & \multicolumn{1}{c|}{1.48}          & 0.51                 & 0.97                 & 1.07                 & \multicolumn{1}{c|}{1.16}          & 0.41                 & 1.05                 & 1.49                 & 1.63                 \\
            \multicolumn{1}{l|}{ConvSeq2Seq\cite{Li_2018_CVPR}}         & 0.51                 & 0.82                 & 1.21                 & \multicolumn{1}{c|}{1.38}          & 0.59                 & 1.13                 & 1.51                 & \multicolumn{1}{c|}{1.65}          & 0.29                 & 0.60                 & 1.12                 & \multicolumn{1}{c|}{1.37}          & 0.63                 & 0.91                 & 1.19                 & \multicolumn{1}{c|}{1.29}          & 0.39                 & 0.61                 & 1.02                 & 1.18                 \\
            \multicolumn{1}{l|}{AGED w/ adv\cite{Gui_2018_ECCV}}         & 0.56                 & 0.81                 & 1.30                 & \multicolumn{1}{c|}{1.46}          & \textbf{0.19}        & \textbf{0.34}        & \textbf{0.50}        & \multicolumn{1}{c|}{\textbf{0.68}} & 0.31                 & 0.58                 & 1.12                 & \multicolumn{1}{c|}{1.34}          & 0.46                 & 0.78                 & 1.01                 & \multicolumn{1}{c|}{1.07}          & 0.41                 & 0.76                 & 1.05                 & 1.19                 \\
            \multicolumn{1}{l|}{SkelNet\cite{Guo_2019_human}}           & 0.50                 & 0.84                 & 1.28                 & \multicolumn{1}{c|}{1.45}          & 0.58                 & 1.12                 & 1.52                 & \multicolumn{1}{c|}{1.64}          & 0.29                 & 0.62                 & 1.19                 & \multicolumn{1}{c|}{1.44}          & 0.58                 & 0.84                 & 1.17                 & \multicolumn{1}{c|}{1.24}          & 0.40                 & 0.61                 & 1.01                 & 1.15                 \\
            \multicolumn{1}{l|}{TD-DCT\cite{Mao_2019_ICCV}}             & 0.36                 & 0.60                 & 0.95                 & \multicolumn{1}{c|}{1.13}          & 0.53                 & 1.02                 & 1.35                 & \multicolumn{1}{c|}{1.48}          & 0.19                 & 0.44                 & 1.01                 & \multicolumn{1}{c|}{1.24}          & 0.43                 & 0.65                 & 1.05                 & \multicolumn{1}{c|}{1.13}          & \textbf{0.29}        & \textbf{0.45}        & \textbf{0.80}        & \textbf{0.97}        \\ \hline
            \multicolumn{1}{l|}{NAT (Ours)}                           & 0.36                 & 0.59                 & 0.93                 & \multicolumn{1}{c|}{1.08}          & 0.55                 & 0.96                 & 1.28                 & \multicolumn{1}{c|}{1.42}          & \textbf{0.18}        & 0.43                 & 0.93                 & \multicolumn{1}{c|}{1.16}          & 0.46                 & 0.67                 & 0.96                 & \multicolumn{1}{c|}{1.03}          & \textbf{0.29}        & 0.46                 & \textbf{0.80}        & 0.98                 \\
            \multicolumn{1}{l|}{mNAT (Ours)}                             & \textbf{0.33}        & \textbf{0.51}        & \textbf{0.79}        & \multicolumn{1}{c|}{\textbf{0.94}} & 0.53                 & 0.92                 & 1.15                 & \multicolumn{1}{c|}{1.28}          & \textbf{0.18}        & \textbf{0.38}        & \textbf{0.81}        & \multicolumn{1}{c|}{\textbf{1.00}} & \textbf{0.40}        & \textbf{0.55}        & \textbf{0.85}        & \multicolumn{1}{c|}{\textbf{0.89}} & \textbf{0.29}        & 0.46                 & 0.84                 & 1.04                 \\ \Xhline{1pt}
            \multicolumn{1}{l|}{}                                       & \multicolumn{4}{c|}{\textbf{Sitting Down}}                                                              & \multicolumn{4}{c|}{\textbf{Taking Photo}}                                                              & \multicolumn{4}{c|}{\textbf{Waiting}}                                                                   & \multicolumn{4}{c|}{\textbf{Walking Dog}}                                                               & \multicolumn{4}{c}{\textbf{Walking Together}}                                             \\
            \multicolumn{1}{l|}{\textbf{milliseconds}}                  & 80                   & 160                  & 320                  & \multicolumn{1}{c|}{400}           & 80                   & 160                  & 320                  & \multicolumn{1}{c|}{400}           & 80                   & 160                  & 320                  & \multicolumn{1}{c|}{400}           & 80                   & 160                  & 320                  & \multicolumn{1}{c|}{400}           & 80                   & 160                  & 320                  & 400                  \\ \Xhline{1pt}
            \multicolumn{1}{l|}{Zero-velocity\cite{Martinez_2017_CVPR}} & 0.39                 & 0.74                 & 1.07                 & \multicolumn{1}{c|}{1.19}          & 0.25                 & 0.51                 & 0.79                 & \multicolumn{1}{c|}{0.92}          & 0.34                 & 0.67                 & 1.22                 & \multicolumn{1}{c|}{1.47}          & 0.60                 & 0.98                 & 1.36                 & \multicolumn{1}{c|}{1.50}          & 0.33                 & 0.66                 & 0.94                 & 0.99                 \\
            \multicolumn{1}{l|}{Res-GRU\cite{Martinez_2017_CVPR}}       & 0.39                 & 0.81                 & 1.40                 & \multicolumn{1}{c|}{1.62}          & 0.24                 & 0.51                 & 0.90                 & \multicolumn{1}{c|}{1.05}          & 0.28                 & 0.53                 & 1.02                 & \multicolumn{1}{c|}{1.14}          & 0.56                 & 0.91                 & 1.26                 & \multicolumn{1}{c|}{1.40}          & 0.31                 & 0.58                 & 0.87                 & 0.91                 \\
            \multicolumn{1}{l|}{ConvSeq2Seq\cite{Li_2018_CVPR}}         & 0.41                 & 0.78                 & 1.16                 & \multicolumn{1}{c|}{1.31}          & 0.23                 & 0.49                 & 0.88                 & \multicolumn{1}{c|}{1.06}          & 0.30                 & 0.62                 & 1.09                 & \multicolumn{1}{c|}{1.30}          & 0.59                 & 1.00                 & 1.32                 & \multicolumn{1}{c|}{1.44}          & 0.27                 & 0.52                 & 0.71                 & 0.74                 \\
            \multicolumn{1}{l|}{AGED w/ adv\cite{Gui_2018_ECCV}}         & 0.33                 & 0.62                 & 0.98                 & \multicolumn{1}{c|}{1.10}          & 0.23                 & 0.48                 & 0.81                 & \multicolumn{1}{c|}{0.95}          & 0.24                 & 0.50                 & 1.02                 & \multicolumn{1}{c|}{1.13}          & 0.50                 & 0.81                 & 1.15                 & \multicolumn{1}{c|}{1.27}          & 0.23                 & 0.41                 & 0.56                 & 0.62                 \\
            \multicolumn{1}{l|}{SkelNet\cite{Guo_2019_human}}           & 0.37                 & 0.72                 & 1.05                 & \multicolumn{1}{c|}{1.17}          & 0.24                 & 0.47                 & 0.78                 & \multicolumn{1}{c|}{0.93}          & 0.30                 & 0.63                 & 1.17                 & \multicolumn{1}{c|}{1.40}          & 0.54                 & 0.88                 & 1.20                 & \multicolumn{1}{c|}{1.35}          & 0.27                 & 0.53                 & 0.68                 & 0.74                 \\
            \multicolumn{1}{l|}{TD-DCT\cite{Mao_2019_ICCV}}             & \textbf{0.30}        & \textbf{0.61}        & \textbf{0.90}        & \multicolumn{1}{c|}{\textbf{1.00}} & \textbf{0.14}        & 0.34                 & 0.58                 & \multicolumn{1}{c|}{0.70}          & 0.23                 & 0.50                 & 0.91                 & \multicolumn{1}{c|}{1.14}          & 0.46                 & 0.79                 & 1.12                 & \multicolumn{1}{c|}{1.29}          & \textbf{0.15}        & 0.34                 & 0.52                 & 0.57                 \\ \hline
            \multicolumn{1}{l|}{NAT (Ours)}                           & 0.31                 & 0.63                 & 0.92                 & \multicolumn{1}{c|}{1.05}          & 0.17                 & 0.37                 & 0.59                 & \multicolumn{1}{c|}{0.71}          & 0.23                 & \textbf{0.48}        & 0.87                 & \multicolumn{1}{c|}{1.07}          & \textbf{0.40}        & \textbf{0.69}        & 1.00                 & \multicolumn{1}{c|}{1.14}          & \textbf{0.15}        & 0.31                 & \textbf{0.45}        & \textbf{0.51}        \\
            \multicolumn{1}{l|}{mNAT (Ours)}                             & 0.34                 & 0.66                 & 0.94                 & \multicolumn{1}{c|}{1.06}          & \textbf{0.14}        & \textbf{0.32}        & \textbf{0.54}        & \multicolumn{1}{c|}{\textbf{0.68}} & \textbf{0.22}        & \textbf{0.48}        & \textbf{0.79}        & \multicolumn{1}{c|}{\textbf{0.97}} & 0.43                 & 0.70                 & \textbf{0.86}        & \multicolumn{1}{c|}{\textbf{1.03}} & 0.16                 & \textbf{0.28}        & 0.51                 & 0.62                 \\ \Xhline{1pt}
                                                                        & \multicolumn{1}{l}{} & \multicolumn{1}{l}{} & \multicolumn{1}{l}{} & \multicolumn{1}{l}{}               & \multicolumn{1}{l}{} & \multicolumn{1}{l}{} & \multicolumn{1}{l}{} & \multicolumn{1}{l}{}               & \multicolumn{1}{l}{} & \multicolumn{1}{l}{} & \multicolumn{1}{l}{} & \multicolumn{1}{l}{}               & \multicolumn{1}{l}{} & \multicolumn{1}{l}{} & \multicolumn{1}{l}{} & \multicolumn{1}{l}{}               & \multicolumn{1}{l}{} & \multicolumn{1}{l}{} & \multicolumn{1}{l}{} & \multicolumn{1}{l}{}
         \end{tabular}
      }
   \end{center}
\end{table*}
   
\section{Experiments}
\label{sec:exper}

In this section, we first introduce two popular motion capture benchmarks: Human3.6M\cite{ionescu2013human3} and CMU Motion Capture dataset\cite{CMU_Motion} (CMU-Mocap) as well as the implementation details and evaluation metrics.
We then demonstrate our results compared with the current state-of-the-arts and ablation studies.

\subsection{Datasets}
   \textbf{Human3.6M}
   The Human3.6M dataset \cite{ionescu2013human3} is the largest publicly available dataset for human motion research so far, which contains 3.6 million 3D poses recorded by Vicon motion capture system.
   It contains 15 activity scenarios including walking, eating, smoking, and discussion. 
   Seven subjects are involved in the dataset, each of which performs two sequences for each action. In total, each sequence contains about 3000 to 5000 frames.
   Each frame consists of 34 rows of data, including a global translation, a global rotation and 32 joint rotations with respect to its parent joint. Each joint is represented as an exponential map (axis-angle) form.
   Following the standard protocol \cite{Martinez_2017_CVPR, Li_2018_CVPR, Pavllo_2018_BMVC}, all sequences are downsampled to a frame rate of 25fps; global translation and global rotation are discarded.
   The Subject 5 (S5) is used in testing while the others are used in training.

   \textbf{CMU Motion Capture}
   The CMU Motion Capture \cite{CMU_Motion} is a large dataset including actions such as walking, running, dancing. 
   Different from Human3.6M, the CMU-Mocap dataset has 38 joints in total. Therefore, it has a different skeleton configuration.
   Li \emph{et al.}\cite{Li_2018_CVPR} first conduct experiments on CMU-Mocap with selected eight actions. 
   We follow their experiment setting with 86293 frames in total.
   Five subjects are used for training while one subject is used for testing.
   Similar to Human3.6M, all sequences are also downsampled to 25fps.
   Global translation and global rotation are discarded.
   
   \begin{table*}[]
      \begin{center}
      \caption{Comparison of Mean Joint Error of angle space between our model and the state-of-the-art methods on 8 actions as well as average result of CMU-Mocap dataset.}
      \label{tab:cmu}
         \resizebox{\textwidth}{!}{
            \begin{tabular}{l|cccc|cccc|cccc|cccc|cccc}
               \Xhline{1pt}
                                                      & \multicolumn{4}{c|}{\textbf{Basketball}}                                                                            & \multicolumn{4}{c|}{\textbf{Basketball Signal}}                                                                     & \multicolumn{4}{c|}{\textbf{Directing Traffic}}                                                            & \multicolumn{4}{c|}{\textbf{Jumping}}                                                                                                 & \multicolumn{4}{c}{\textbf{Running}}                                                                                        \\
               \textbf{milliseconds}                  & 80                                & 160                      & 320                      & 400                       & 80                       & 160                      & 320                      & 400                                & 80                       & 160                      & 320                      & 400                       & 80                                & 160                               & 320                               & 400                       & 80                       & 160                      & 320                               & 400                               \\ \Xhline{1pt}
               Zero-velocity\cite{Martinez_2017_CVPR} & 0.48                              & 0.82                     & 1.40                     & 1.64                      & 0.24                     & 0.44                     & 0.75                     & 0.86                               & 0.30                     & 0.57                     & 0.90                     & 1.01                      & 0.36                              & 0.62                              & 1.48                              & 1.68                      & 0.56                     & 1.00                     & 1.38                              & 1.46                              \\
               Res-GRU\cite{Martinez_2017_CVPR}       & 0.50                              & 0.80                     & 1.27                     & 1.45                      & 0.41                     & 0.76                     & 1.32                     & 1.54                               & 0.33                     & 0.59                     & 0.93                     & 1.10                      & 0.56                              & 0.88                              & 1.77                              & 2.02                      & 0.33                     & 0.50                     & 0.66                              & 0.75                              \\
               ConvSeq2Seq\cite{Li_2018_CVPR}         & 0.37                              & 0.62                     & 1.07                     & 1.18                      & 0.32                     & 0.59                     & 1.04                     & 1.24                               & 0.25                     & 0.56                     & 0.89                     & 1.00                      & 0.39                              & 0.60                              & 1.36                              & 1.56                      & 0.28                     & \textbf{0.41}            & \textbf{0.52}                     & 0.57                              \\ \hline
               NAT (Ours)                           & \multicolumn{1}{l}{\textbf{0.34}} & \multicolumn{1}{l}{0.52} & \multicolumn{1}{l}{0.88} & \multicolumn{1}{l|}{1.03} & \multicolumn{1}{l}{0.19} & \multicolumn{1}{l}{0.28} & \multicolumn{1}{l}{0.49} & \multicolumn{1}{l|}{\textbf{0.61}} & \multicolumn{1}{l}{0.22} & \multicolumn{1}{l}{0.44} & \multicolumn{1}{l}{0.67} & \multicolumn{1}{l|}{0.79} & \multicolumn{1}{l}{\textbf{0.38}} & \multicolumn{1}{l}{\textbf{0.56}} & \multicolumn{1}{l}{\textbf{1.27}} & \multicolumn{1}{l|}{1.47} & \multicolumn{1}{l}{0.26} & \multicolumn{1}{l}{0.49} & \multicolumn{1}{l}{\textbf{0.52}} & \multicolumn{1}{l}{\textbf{0.56}} \\
               mNAT (Ours)                             & \textbf{0.34}                     & \textbf{0.49}            & \textbf{0.86}            & \textbf{1.01}             & \textbf{0.15}            & \textbf{0.24}            & \textbf{0.48}            & \textbf{0.61}                      & \textbf{0.20}            & \textbf{0.41}            & \textbf{0.65}            & \textbf{0.77}             & \textbf{0.38}                     & \textbf{0.56}                     & 1.29                              & \textbf{1.45}             & \textbf{0.24}            & 0.43                     & 0.53                              & \textbf{0.56}                     \\ \Xhline{1pt}
                                                      & \multicolumn{4}{c|}{\textbf{Soccer}}                                                                                & \multicolumn{4}{c|}{\textbf{Walking}}                                                                               & \multicolumn{4}{c|}{\textbf{Wash Window}}                                                                  & \multicolumn{4}{c|}{\textbf{Average}}                                                                                                 &                          &                          &                                   &                                   \\
               \textbf{milliseconds}                  & 80                                & 160                      & 320                      & 400                       & 80                       & 160                      & 320                      & 400                                & 80                       & 160                      & 320                      & 400                       & 80                                & 160                               & 320                               & 400                       &                          &                          &                                   &                                   \\ \Xcline{1-17}{1pt}
               Zero-velocity\cite{Martinez_2017_CVPR} & 0.27                              & 0.48                     & 0.92                     & 1.10                      & 0.41                     & 0.60                     & 0.83                     & 0.95                               & 0.34                     & 0.57                     & 0.90                     & 1.10                      & 0.37                              & 0.64                              & 1.07                              & 1.22                      &                          &                          &                                   &                                   \\
               Res-GRU\cite{Martinez_2017_CVPR}       & 0.29                              & 0.51                     & 0.88                     & 0.99                      & 0.35                     & 0.47                     & 0.60                     & 0.65                               & 0.30                     & 0.46                     & 0.72                     & 0.91                      & 0.38                              & 0.62                              & 1.02                              & 1.18                      &                          &                          &                                   &                                   \\
               ConvSeq2Seq\cite{Li_2018_CVPR}         & 0.26                              & 0.44                     & 0.75                     & 0.87                      & 0.35                     & 0.44                     & 0.45                     & 0.50                               & 0.30                     & 0.47                     & 0.80                     & 1.01                      & 0.32                              & 0.52                              & 0.86                              & 0.99                      &                          &                          &                                   &                                   \\ \cline{1-17}
               NAT (Ours)                           & \multicolumn{1}{l}{0.23}          & \multicolumn{1}{l}{0.34} & \multicolumn{1}{l}{0.61} & \multicolumn{1}{l|}{0.73} & \multicolumn{1}{l}{0.33} & \multicolumn{1}{l}{0.39} & \multicolumn{1}{l}{0.42} & \multicolumn{1}{l|}{0.48}          & \multicolumn{1}{l}{0.27} & \multicolumn{1}{l}{0.40} & \multicolumn{1}{l}{0.72} & \multicolumn{1}{l|}{0.93} & \multicolumn{1}{l}{0.28}          & \multicolumn{1}{l}{0.43}          & \multicolumn{1}{l}{0.70}          & \multicolumn{1}{l|}{0.83} & \multicolumn{1}{l}{}     & \multicolumn{1}{l}{}     & \multicolumn{1}{l}{}              & \multicolumn{1}{l}{}              \\
               mNAT (Ours)                             & \textbf{0.20}                     & \textbf{0.33}            & \textbf{0.59}            & \textbf{0.72}             & \textbf{0.31}            & \textbf{0.37}            & \textbf{0.40}            & \textbf{0.46}                      & \textbf{0.23}            & \textbf{0.36}            & \textbf{0.66}            & \textbf{0.86}             & \textbf{0.26}                     & \textbf{0.40}                     & \textbf{0.68}                     & \textbf{0.80}             &                          &                          &                                   &                                   \\ \Xhline{1pt}
            \end{tabular}
         }
      \end{center}
   \end{table*}

\subsection{Implementation Details}
   \textbf{Network and Training Details}
   Our model is based on QuaterNet\cite{Pavllo_2018_BMVC} and quaternion is used as input representation of joints.
   Similar to the previous work \cite{Martinez_2017_CVPR,Li_2018_CVPR,Pavllo_2018_BMVC}, our model is also trained on all actions.
   Both context encoder and frame decoder stack 6 GCN-TCN building blocks with residual connections.
   When the dimension changes, an extra 1D Conv is performed to transform dimension.
   The number of channels in the context encoder is 64, 64, 128, 128, 256, 256 for each building block, respectively. 
   The number of channels in the frame decoder is 256, 128, 128, 64, 64, 4, respectively.
   We use a pre-defined graph with relation of joints for GCN based on specific dataset.
   The kernel size $ks=9$ for TCN in context encoder while $ks=1$ in frame decoder.
   Each TCN is performed with the same padding mode to ensure that the temporal size stays unchanged.
   LeakyReLU is utilized as the non-linear activation with a rate of 0.01.
   We apply $\alpha=10$ and $\beta=500$ for positional encoding as this setting achieves optimal performance in our observation.
   The dropout rate is set to 0.5 for ARC module.
   The whole model is lightweight with 2.28 MB.

   For both datasets, ADAM\cite{kingma2014adam} is selected as the optimizer in our experiment. The initial learning rate is 0.001 with a 0.9995 decay in every epoch.
   The gradient clip norm is set to 0.1 and the mini-batch is composed of 60 samples.
   Following the previous work\cite{Pavllo_2018_BMVC}, $\lambda_{\mathrm{pnlty}}$ is set to 0.01.
   Also, $\lambda_{\mathrm{cls}}$ is set to 0.01 to balance two tasks. We conduct an ablation study on the effects of $\lambda_{\mathrm{cls}}$ on final results.
   Our model is trained with PyTorch\cite{paszke2019pytorch} framework for 3000 epochs on a single NVIDIA 1080TI GPU.

   \textbf{Evaluation Metrics and Baselines}
   For Human3.6M, we report our results on both short-term ($80 \sim 400$ ms) and long-term ($80 \sim 1000$ ms). 
   For CMU-Mocap, due to the space limit, we only report short-term results. 
   For all datasets, 50 frames ($2000$ ms) are given. Following the previous work\cite{Martinez_2017_CVPR}, the error is measured as the average Euclidean distance between predicted joints and ground truth in Euler angle space.

   To evaluate the performance of our model, we compare it with five state-of-the-art human motion prediction approaches, namely, Res-GRU\cite{Martinez_2017_CVPR}, ConvSeq2Seq\cite{Li_2018_CVPR}, AGED (w/ adv)\cite{Gui_2018_ECCV}, SkelNet\cite{Guo_2019_human}, TD-DCT\cite{Mao_2019_ICCV}, as well as one baseline method Zero-velocity\cite{Martinez_2017_CVPR}.
   Note that the results of SkelNet is based on their open source project\footnote{\url{https://github.com/CHELSEA234/SkelNet_motion_prediction}} since they do not provide results for all 15 actions.
   All the other results are referred from their original papers.

   \begin{table}
      \begin{center}
      \caption{Comparison of mean joint error of angle space for long-term of Human3.6M dataset}
      \label{tab:long-term}
         \resizebox{0.48\textwidth}{!}{
            \begin{tabular}{l|cc|cc|cc|cc}
            \Xhline{1pt}
                                                   & \multicolumn{2}{c|}{\textbf{Walking}} & \multicolumn{2}{c|}{\textbf{Eating}} & \multicolumn{2}{c|}{\textbf{Smoking}} & \multicolumn{2}{c}{\textbf{Discussion}} \\
            \textbf{milliseconds}                  & 560               & 1000              & 560               & 1000             & 560               & 1000              & 560                & 1000               \\ \Xhline{1pt}
            Zero-velocity\cite{Martinez_2017_CVPR} & 1.35              & 1.32              & 1.04              & 1.38             & 1.02              & 1.69              & 1.41               & 1.96               \\
            ERD \cite{Fragkiadaki_2015_ICCV}       & 2.00              & 2.38              & 2.36              & 2.41             & 3.68              & 3.82              & 3.47               & 2.92               \\
            SRNN \cite{Jain_2016_CVPR}             & 1.81              & 2.20              & 2.49              & 2.82             & 3.24              & 2.42              & 2.48               & 2.93               \\ \hline
            Res-GRU\cite{Martinez_2017_CVPR}       & 0.93              & 1.03              & 0.95              & 1.08             & 1.25              & 1.50              & 1.43               & 1.69               \\
            ConvSeq2Seq\cite{Li_2018_CVPR}         & 0.86              & 0.92              & 0.89              & 1.24             & 0.97              & 1.62              & 1.44               & 1.86               \\
            AGED w/ adv \cite{Gui_2018_ECCV}       & 0.78              & 0.91              & 0.86              & 0.93             & 1.06              & \textbf{1.21}     & 1.25               & 1.30               \\
            SkelNet \cite{Guo_2019_human}          & 0.79              & 0.83              & 0.84              & 1.06             & 0.98              & \textbf{1.21}     & 1.39               & 1.75               \\ \hline
            NAT (Ours)                           & 0.56              & 0.58              & 0.71              & 0.96             & 0.78              & 1.39              & \textbf{1.13}      & 1.35               \\
            mNAT (Ours)                            & \textbf{0.54}     & \textbf{0.50}     & \textbf{0.64}     & \textbf{0.87}    & \textbf{0.73}     & 1.26              & 1.18               & \textbf{1.22}      \\ \Xhline{1pt}
            \end{tabular}
         }
      \end{center}
   \end{table}

   \begin{figure}[!t]
      \centering
      \includegraphics[width=1\linewidth]{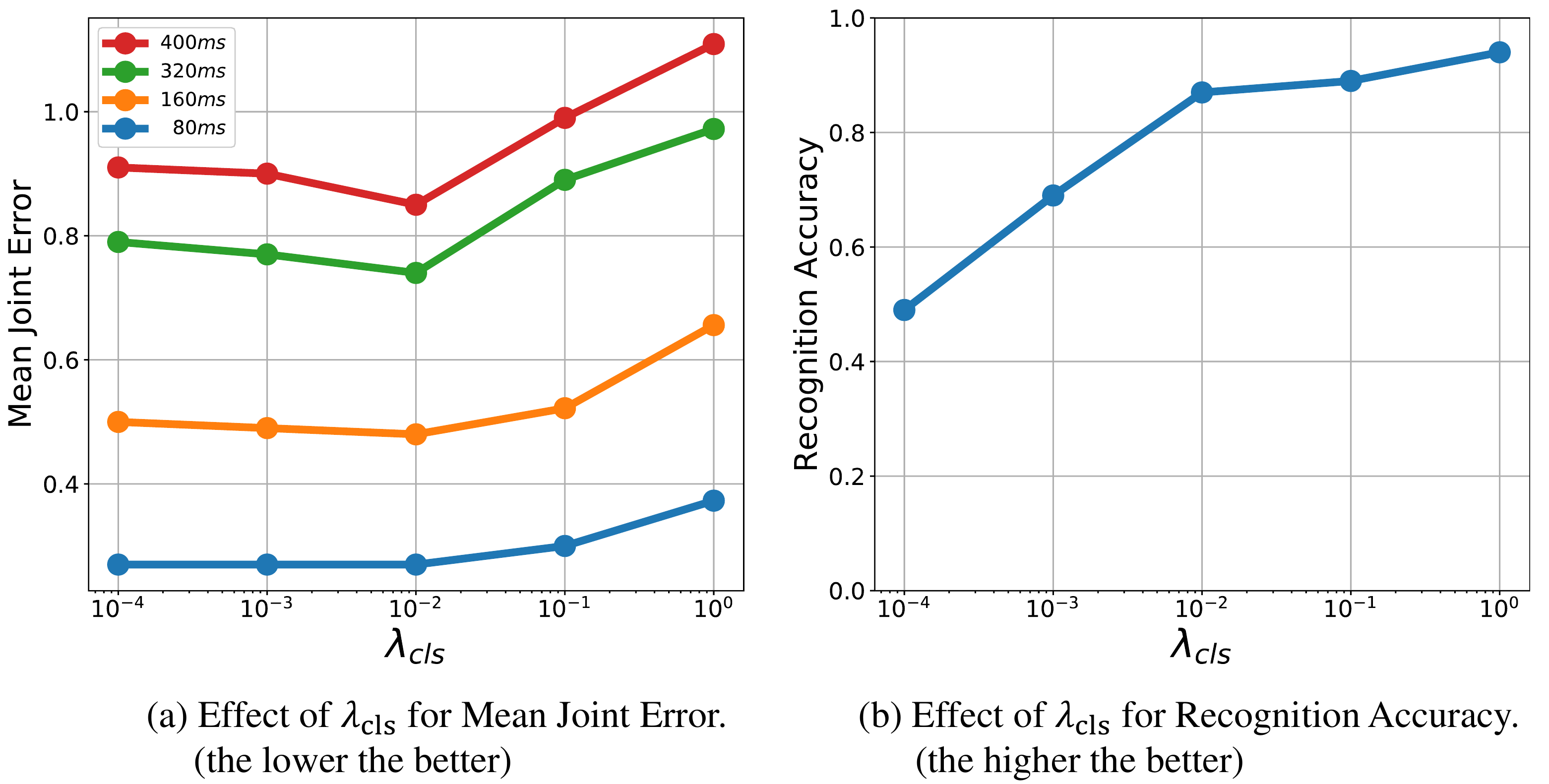}
      \caption{
         Illustration of effects of $\lambda_\mathrm{cls}$ for both mean joint error and recognition accuracy of Human3.6M dataset.
         }
      \label{fig:lambda}
   \end{figure}

\begin{figure*}[!t]
   \centering
     \includegraphics[width=0.95\linewidth]{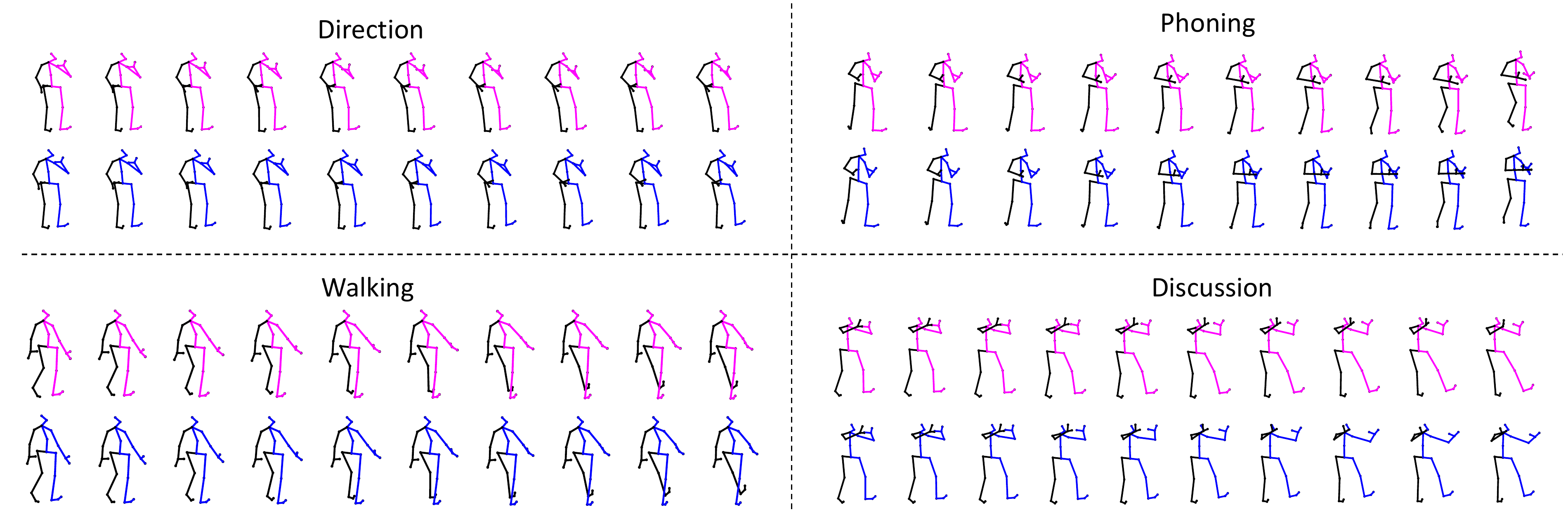}
     \caption{
        Qualitative results based on Human3.6M dataset.
        Starting from top left, we demonstrate for four actions: direction, phoning, walking, and discussion.
        For each action, the top (in red color) and the bottom (in blue color) are the ground truth and our prediction respectively.
      }
   \label{fig:vis}
\end{figure*}

\subsection{Comparison with State-of-the-Art}
   We evaluate our model on two popular datasets, Human3.6M and CMU-Mocap.
   We post results for NAT and the NAT variant which is equipped with multitask training paradigm (mNAT).
   Table \ref{tab:avg} reports for the average of the mean joint error of angle space of Human3.6M dataset for all 15 human actions.
   With the help of our non-autoregressive setting, both our NAT and mNAT outperform all the state-of-the-art approaches in average of 15 actions.
   Note that although our results keep the same level in 80 and 160 ms, with the latest state-of-the-art (TD-DCT \cite{Mao_2019_ICCV}), which almost approaches the upper limit.
   Our methods surpass a large margin (0.09 and 0.10) in 320ms and 400ms, respectively.
   Note that with the help of multitask learning, the average performance gains 0.02, 0.05, 0.06 for 160, 320, and 400ms compared with NAT.
   The interesting part lies in that the ARC module helps to improve performance in the long run.
   This is because, in the short term, the class information is not important since the inertance is the key factor.
   However, in the long term, it is not easy to predict since much more factors are taken into consideration.
   Therefore, to predict target pose and the action label simultaneously helps a lot by introducing the guidance from activity, leading to convincing performance gain. 

   Table \ref{tab:all} shows the results for all 15 human actions in detail. NAT and mNAT also achieve the best performance on most actions.
   Of all the action categories, we also notice the abnormal performance in action ``Phoning''.
   After examination, we finally realize that the high error is due to the \textbf{discontinuity} of the test data.
   We observe that the finger joint data are abnormal, which largely affects the performance of the mean joint error.
   Since this is an end-effector joint with little influence on others, the performance in 3D space is not affected, which is also discussed in \cite{Mao_2019_ICCV}.
   As for the results of other state-of-the-art approaches, we faithfully report their performances from their original papers.

   In Table \ref{tab:long-term}, we also report the performance of long-term of Human3.6M dataset.
   We observe an obvious improvement for actions such as ``Walking'', ``Eating''.
   For example, the performance gain reaches 0.25 and 0.33 for 560 and 1000 ms of ``Walking''.
   This proves that the inertance still exists in long term such that the non-autoregressive model could be utilized.
   The results also show the existence of error accumulation of the previous RNN-based autoregressive methods, in which the mean joint error grows fast with the evolution of time.

   Table \ref{tab:cmu} reports the results on the CMU-Mocap dataset of the mean joint error. Our model also achieves the best performance on all 8 actions and the lower average error than the previous baselines, which verifies what we discuss above.

   We also present the qualitative results for Human 3.6M dataset in Fig. \ref{fig:vis}.
   From the visualization, our prediction is quite close to the ground truth.
   Note that even for the ``phoning'' action, our model still gives plausible and reliable results.

\subsection{Ablation Study}
\textbf{Balance of losses.} In this model, we present the ARC module which makes use of multitask learning, leading to convincing improvement.
However, the performance gains for prediction and classification are traded off by the hyper-parameter $\lambda_{\mathrm{cls}}$.
We study the effects of $\lambda_{\mathrm{cls}}$ on two tasks.

\begin{table}[]
   \centering
   \caption{The average mean joint error of different settings of $\alpha$, $\beta$ of Human3.6M dataset.}
   \label{tab:alpha_beta}
   \resizebox{0.34\textwidth}{!}{
      \begin{tabular}{cc|llll}
         \Xhline{1pt}
         \multicolumn{1}{l}{} & \multicolumn{1}{l|}{} & \multicolumn{4}{c}{\textbf{Average (ms)}}                                                            \\
         \textbf{$\alpha$}    & \textbf{$\beta$}      & \multicolumn{1}{c}{80} & \multicolumn{1}{c}{160} & \multicolumn{1}{c}{320} & \multicolumn{1}{c}{400} \\ \Xhline{1pt}
         1                    & 10000                 & 0.28                   & 0.51                    & 0.84                    & 0.96                    \\ \hline
         0.1                  & 10000                 & 0.28                   & 0.52                    & 0.84                    & 0.94                    \\
         10                   & 10000                 & 0.28                   & 0.50                    & 0.80                    & 0.92                    \\ \hline
         1                    & 500                   & 0.28                   & 0.51                    & 0.82                    & 0.94                    \\
         1                    & 20000                 & 0.29                   & 0.54                    & 0.85                    & 0.99                    \\ \hline
         10                   & 500                   & \textbf{0.27}          & \textbf{0.50}           & \textbf{0.79}           & \textbf{0.91}                    \\ \Xhline{1pt}
         \end{tabular}
   }
\end{table}

We evaluate our model and present the mean joint error and recognition accuracy for different $\lambda_{\mathrm{cls}}$.
Fig. \ref{fig:lambda} illustrates the performances for both tasks.
From the result, we observe that: 
1) The recognition accuracy grows with the improvement of $\lambda_{\mathrm{cls}}$. 
2) For human motion prediction task, the performance is robust when $\lambda_{\mathrm{cls}}$ is low. 
However, an obvious performance drop is observed when $\lambda_{\mathrm{cls}}$ gets larger. 
This is mainly because motion prediction is more difficult than classification task. 
When we pay too much attention to classification, the model becomes unbalanced.
Therefore, we choose $\lambda_{\mathrm{cls}}=0.01$ in our paper.

\begin{table}[]
   \centering
   \caption{The average mean joint error of different settings of kernel size in TCN of Human3.6M dataset.}
   \label{tab:kernel_size}
   \resizebox{0.34\textwidth}{!}{
      \begin{tabular}{c|llll}
         \Xhline{1pt}
         \multicolumn{1}{l|}{} & \multicolumn{4}{c}{\textbf{Average (ms)}}                                                            \\
         kernel size           & \multicolumn{1}{c}{80} & \multicolumn{1}{c}{160} & \multicolumn{1}{c}{320} & \multicolumn{1}{c}{400} \\ \Xhline{1pt}
         3                     & 0.40                   & 0.70                    & 1.06                    & 1.19                    \\
         5                     & 0.40                   & 0.71                    & 1.07                    & 1.20                    \\
         7                     & 0.40                   & 0.67                    & 0.98                    & 1.11                    \\ \hline
         9                     & \textbf{0.27}          & \textbf{0.50}           & \textbf{0.79}           & 0.91                    \\
         11                    & 0.28                   & \textbf{0.50}           & 0.81                    & \textbf{0.90}           \\ \Xhline{1pt}
         \end{tabular}
   }
\end{table}

\begin{table}[]
   \centering
   \caption{The average mean joint error of different settings of graph type in GCN of Human3.6M dataset.}
   \label{tab:graph_type}
   \resizebox{0.34\textwidth}{!}{
      \begin{tabular}{c|llll}
         \Xhline{1pt}
         \multicolumn{1}{l|}{} & \multicolumn{4}{c}{\textbf{Average (ms)}}                                                            \\
         graph type            & \multicolumn{1}{c}{80} & \multicolumn{1}{c}{160} & \multicolumn{1}{c}{320} & \multicolumn{1}{c}{400} \\ \Xhline{1pt}
         no graph              & 0.28                   & 0.53                    & 0.85                    & 0.98                    \\
         random graph          & 0.29                   & 0.52                    & 0.82                    & 0.96                    \\ \hline
         forward               & \textbf{0.27}          & 0.51                    & 0.82                    & 0.93                    \\
         backward              & 0.28                   & 0.52                    & 0.83                    & 0.95                    \\
         bi-directional        & \textbf{0.27}          & \textbf{0.50}           & \textbf{0.79}           & \textbf{0.91}           \\ \Xhline{1pt}
         \end{tabular}
   }
\end{table}

\textbf{Effects of $\alpha$ and $\beta$.} 
In Sec \ref{sec:nat}, we discuss that the original setting of positional encoding module might be sub-optimal due to the domain gap between NMT and motion prediction.
In Table \ref{tab:alpha_beta}, we study the influence of different settings of $\alpha$ and $\beta$.
From the results, we observe that the performance gains with the increase of $\alpha$ and the decrease of $\beta$.
We explain this observation with the function of two parameters.
On one hand, $\alpha$ controls the magnitude of time index $t$.
When $\alpha$ becomes larger, the positional embedding vectors become more distinguishable, leading to a high-quality prediction.
On the other hand, $\beta$ controls the frequency of each dimension.
As can be easily observed from Fig. \ref{fig:pos_emb}, with the high value of $\beta$, lots of dimensions are wasted due to the limited number of poses to be predicted.
In conclusion, our ablation study shows that a large value of $\alpha$ and a small value of $\beta$ lead to an optimal performance.

\textbf{Effects of kernel size for TCN.}
We study the effects of kernel size for TCN in this section.
In Table \ref{tab:kernel_size}, we evaluate the mean joint error for different kernel sizes ranging from 3 to 11.
From the results, we find an obvious boundary between 7 and 9. When the kernel size is less than 7, the performance becomes poor. However, limited improvement is found when we further enlarge the kernel size.
Since multiple GCN-TCN blocks are stacked in our model, the choice of kernel size has a direct influence for the receptive field.
Therefore, the receptive field leads to the above observation.
On one hand, when the kernel size is less than 7, the overall receptive field cannot cover the whole skeleton sequence. Thus, the performance drops due to the large information loss.
On the other hand, all the frames are average pooled in the last, simply increasing the kernel size would not make an obvious difference.
In conclusion, we choose to use kernel size 9 due to the computational efficiency.

\textbf{Effects of graph type for GCN.}
We also study the effects of multiple graphs on GCN.
From Table \ref{tab:graph_type}, ``no graph'' denotes that each joint is connected with itself only. 
In practice, we replace the adjacency matrix with an identity matrix $I$.
Similarly, ``random graph'' means that the connection of joints is random.
From the result, both ``no graph'' and ``random graph'' are slightly worse than GCN with people skeleton graph.

Further, we also explore the effects of the graph direction.
On one hand, ``forward'' denotes that the connection of joints is from the central joint to all its end-effectors.
On the other hand, ``backward'' denotes a converse direction.
From the result, GCN with bi-directional graph obtains a better performance, which means that both parent joint and child joint are equally important in human motion prediction task.

\section{Conclusion}
In this paper, we present a novel human motion prediction framework based on a non-autoregressive method.
The framework takes an encoder-decoder model where a simple yet effective non-autoregressive pipeline is adopted in decoding stage while multiple GCN-TCN blocks are performed so as to fully explore the spatio-temporal relation.
In addition, we also find that by predicting human action category, the prediction becomes more feasible and reliable.
In experiments, our approach surpasses all the recent state-of-the-art human motion forecasting methods.

\ifCLASSOPTIONcaptionsoff
  \newpage
\fi



%


\balance
\bibliographystyle{IEEEtran}
\bibliography{IEEEexample}

\end{document}